\documentclass[11pt]{article}

\usepackage[final]{acl}

\usepackage{xcolor}
\usepackage{booktabs}
\usepackage[inline]{enumitem}
\usepackage{multirow}

\usepackage{xcolor}
\definecolor{steelblue}{RGB}{70, 130, 180}
\definecolor{tomato}{RGB}{255, 99, 71}
\definecolor{seagreen}{RGB}{46, 139, 87}

\usepackage{times}
\usepackage{latexsym}

\usepackage[T1]{fontenc}

\usepackage[utf8]{inputenc}

\usepackage{microtype}

\usepackage{inconsolata}

\usepackage{graphicx}
\usepackage{amsfonts}

\title{Parts-of-Speech as Emergent Categories in SAE Latent Space}

\author{
 \textbf{Alessandro Bondielli\textsuperscript{1,2,*}},
 \textbf{Lucia Passaro\textsuperscript{1,2,*}},
 \textbf{Serena Auriemma\textsuperscript{1}},
 \textbf{Alessandro Lenci\textsuperscript{1}}
\\
\\
 \textsuperscript{1}CoLingLab, Department of Philology, Literature and Linguistics, University of Pisa\\
 \textsuperscript{2}Department of Computer Science, University of Pisa\\
\\
 \small{
   *Equal contribution. \textbf{Correspondence:} \href{mailto:alessandro.bondielli@unipi.it}{alessandro.bondielli@unipi.it}, \href{mailto:lucia.passaro@unipi.it}{lucia.passaro@unipi.it}},\\
   \textbf{\textcolor{red}{Preprint version of the paper in the Proceedings of EMNLP 2026.}}
}

\begin{document}
\maketitle
\begin{abstract}
Sparse AutoEncoders (\textsc{SAE}s) offer a promising way to inspect language model representations, but it is still unclear what kind of linguistic structure their latents expose. We use part-of-speech (\textsc{PoS}) categories as a controlled test case to study whether morpho-syntactic information is encoded by individual latents or by structured groups of features. We find that \textsc{PoS} distinctions are highly recoverable from \textsc{SAE} activations, but do not align with one-to-one latent / category mappings. This recoverability is not reducible to lexical memorisation, and Open and Closed \textsc{PoS} classes differ substantially. Categories are supported by compact groups of sparse latents, with substantial variation across tags. These groups remain stable on held-out data, while also showing overlap between related categories. Our results show that SAEs localise morpho-syntactic information in a distributed and category-dependent form rather than through atomic grammatical features.\footnote{Code and Data available here: \url{https://github.com/colinglab/pos-sae-latents}}.
\end{abstract}

\section{Introduction}\label{sec:intro}

Large language models (LLMs) encode a wide range of linguistic regularities in their internal representations, from lexical and syntactic information to more abstract semantic and discourse-level properties. Yet, despite substantial progress in probing and representation analysis, it remains unclear how such information is organized internally, for instance whether linguistic categories correspond to localized and interpretable units, or  they are instead distributed across many dimensions of the representation space \cite{elhage-2022-toy}. This question has become particularly relevant with the growing use of Sparse AutoEncoders (\textsc{SAE}s) as tools for interpreting LLMs \cite{bricken-2023-monosemanticity,cunningham-2024-sparse,templeton-2024-scaling}.

\textsc{SAE}s aim to decompose dense model activations into high-dimensional sparse representations, where individual dimensions (aka \textbf{latents}) are expected to capture more interpretable directions of variation. \textsc{SAE} latents are expected to provide a bridge between low-level model activations and human-interpretable features. This has motivated their use in mechanistic interpretability, where they are often discussed in terms of feature discovery and monosemanticity \cite{elhage-2022-toy,bricken-2023-monosemanticity,templeton-2024-scaling}.

However, the relationship between SAE latents and \textbf{linguistic categories} is still not clear. In fact, the latter result from the combinations of multiple lexical, morphological, syntactic, and distributional features, which need not correspond to individual isolated latents. Understanding whether linguistic abstractions are localized or distributed in \textsc{SAE} spaces is thus important for evaluating what kind of interpretability \textsc{SAE}s provide \cite{kantamneni-2025-sparse,karvonen-2025-saebench,engels-2024-linear}.

\begin{figure}[t]
    \centering
    \includegraphics[width=\columnwidth]{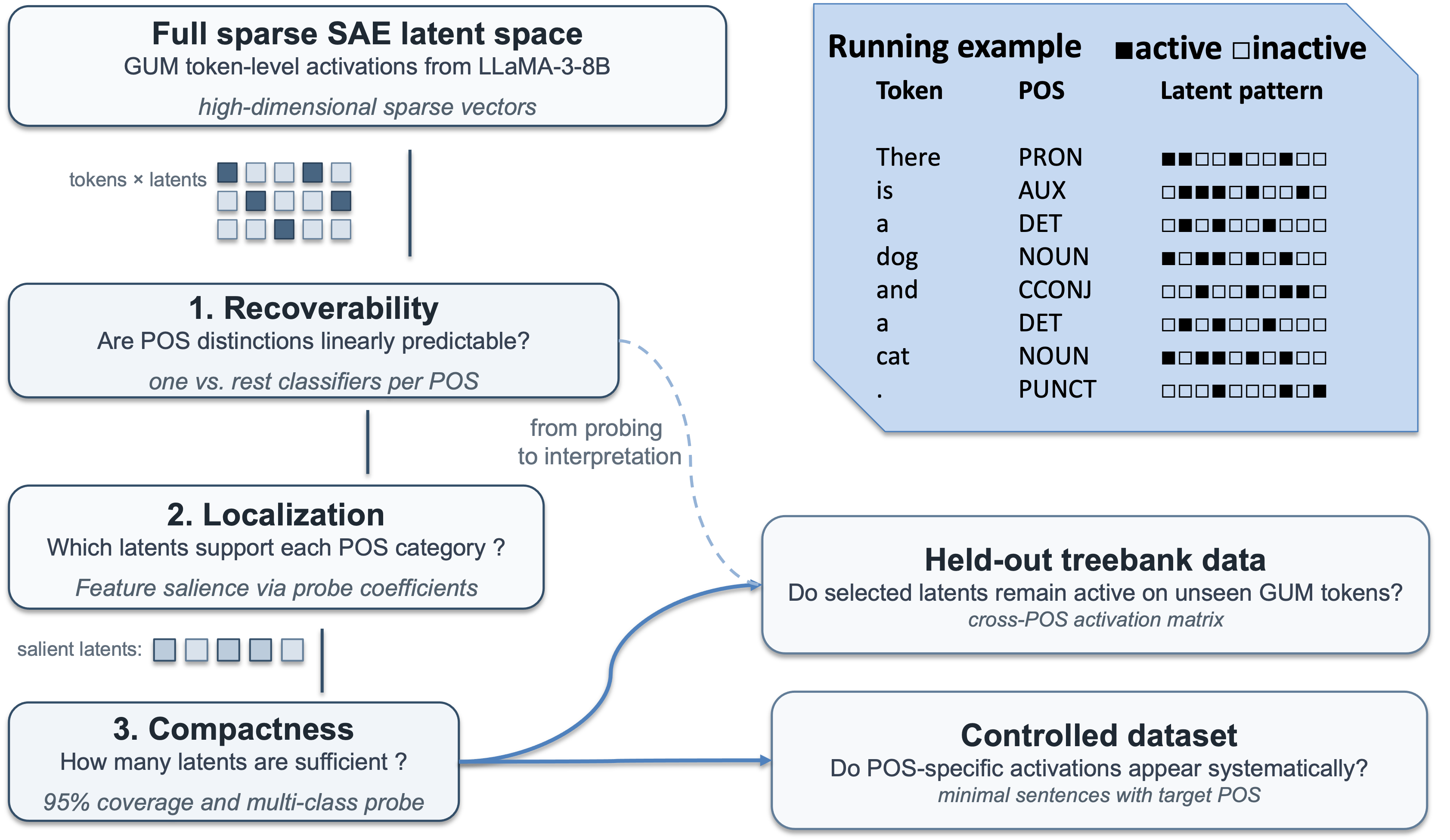}
    \caption{Overview of the workflow. We test \textsc{PoS} recoverability from token-level \textsc{SAE} activations, identify and compact \textsc{PoS}-relevant latent groups, and validate them on held-out and controlled data.}    \label{fig:fig1}
\vspace{-0.1in}

\end{figure}

In this paper, we study this question by targeting part-of-speech (\textsc{PoS}) categories. \textsc{PoS} tags offer a controlled testbed for analyzing morpho-syntactic abstraction: They are discrete, independently annotated, and linguistically interpretable, while also differing in frequency, lexical openness, and syntactic function. For instance, open-class categories such as nouns and verbs are lexically productive and highly variable, whereas closed-class categories such as determiners, conjunctions, and pronouns are more restricted and often tied to specific syntactic roles. This makes \textsc{PoS} a useful setting for testing whether \textsc{SAE} latents behave as localized linguistic features or instead participate in broader distributed representations.

We analyze \textsc{SAE} activations extracted from LLaMA-3-8B \cite{dubey-2024-llama3} on the GUM Corpus Treebank \cite{Zeldes2017}. For each token, we encode its \textsc{SAE} representation into a sparse activation vector and study how gold Universal Dependencies (\textsc{UD}) \textsc{PoS} tags are represented in this latent space. Our experimental design follows a three-step interpretability pipeline: i.) we use \textbf{binary probing classifiers} to test whether individual \textsc{PoS} distinctions are recoverable from \textsc{SAE} activations \cite{belinkov-2022-probing}; ii.)  we use \textbf{feature-salience analysis} to rank the latents most relevant to each \textsc{PoS} category, and \textbf{coverage analysis} to estimate how many of these latents are needed to account for most instances of the category; iii.) we validate the selected latent groups on \textbf{held-out and controlled data}, and test whether their union is sufficient to train a multi-class \textsc{PoS} classifier.

This setup allows us to move beyond standard probing accuracy. A high probing score may show that \textsc{PoS} information is present in \textsc{SAE} activations, but it does not explain how this information is organized \cite{hewitt-liang-2019-designing,pimentel-etal-2020-information,belinkov-2022-probing}. By combining probing, salience, coverage, compact-feature classification, and held-out validation, we can understand whether \textsc{PoS} categories are associated with individual monosemantic latents or with structured groups of sparse latents. We also examine whether different \textsc{PoS} categories are represented by different numbers of latents, which might suggest that the \textsc{SAE} representation reflects differences in the linguistic nature of the categories themselves.

We address three research questions:
\begin{enumerate*}[label=(\roman*)]
    \item \textbf{RQ1:} \textit{Are PoS categories explicitly encoded in \textsc{SAE} latent activations?}
    \item \textbf{RQ2:} \textit{What is the organization of the \textsc{PoS} categories encoding in the \textsc{SAE} latent spaces}?
    \item \textbf{RQ3:} \textit{How stable and systematic are these latent representations across linguistic categories, datasets, and evaluation settings?}
\end{enumerate*}

Our study makes two main contributions. First, we show that \textbf{\textsc{PoS} categories are aligned with structured groups of sparse features}. Through feature-salience and coverage analyses, we quantify the size and organization of these groups, show that it varies substantially across categories, and highlight differences between Open- and Closed-class \textsc{PoS} classes. Second, we show that \textbf{these latent groups are compact yet effective}: Their union preserves strong multi-class \textsc{PoS} classification performance, and they remain stable on held-out data, while still exhibiting overlap across related categories. 

\section{Related Work}

Probing classifiers have long been used to test what linguistic information neural language models encode in their representations \citep{conneau-etal-2018-cram, belinkov-2022-probing}. Prior work shows that lower layers capture morpho-syntactic information such as \textsc{PoS}, while higher layers encode more abstract semantic and discourse properties \citep{tenney-etal-2019-bert, tenney-etal-2019-what, hewitt-manning-2019-structural,rogers-etal-2020-primer}. However, probing accuracy alone is limited: control tasks \citep{hewitt-liang-2019-designing} and information-theoretic critiques \citep{pimentel-etal-2020-information} show that probes can fit arbitrary mappings, and that recoverability does not imply use. We share this concern, but shift the focus from \emph{what} information is present to \emph{how} it is organised at the level of individual sparse latents.

A growing body of work studies mechanistic interpretability in LLMs \cite{sharkey2025open}. Within this area, \textsc{SAE}s map dense activations to high-dimensional sparse vectors whose units are intended to be more monosemantic and interpretable \citep{bricken-2023-monosemanticity, cunningham-2024-sparse}. Subsequent work has improved \textsc{SAE} training through scaling \citep{templeton-2024-scaling} and TopK activations \citep{gao2025scaling}, and released open \textsc{SAE} suites for widely used base models \citep{lieberum-2024-gemma,he-2024-llama}. These studies often identify latents aligned with intuitive concepts, using top-activating examples or automated natural-language explanations, but provide limited evidence on how \emph{theoretically motivated linguistic categories} are represented in latent space.

Recent work also questions whether \textsc{SAE} latents behave as genuinely monosemantic features. \citet{kantamneni-2025-sparse} find that probes trained on \textsc{SAE} latents do not consistently outperform simple baselines across $113$ binary classification tasks, while SAEBench \citep{karvonen-2025-saebench} shows that gains on standard \textsc{SAE} proxy metrics often do not transfer to downstream performance. Still, \textsc{SAE}s remain useful tools for probing LM knowledge and behaviour \cite{duprelatour2025sparseautoencoderlatentattribution, frasertaliente2026nla}.

Closer to our work, \citet{marks-2024-sparse} use \textsc{SAE} features to construct interpretable causal circuits for syntactic phenomena such as subject--verb agreement, suggesting that morpho-syntactic information is at least partly recoverable from \textsc{SAE} space. At the same time, \citet{engels-2024-linear} show that not all language model features are well captured by single linear directions. Existing linguistic analyses of \textsc{SAE}s have mainly focused on isolated phenomena, such as subject--verb agreement, or broad properties such as language identity. We address the open question of whether and how classical morpho-syntactic categories are encoded by latents, using \textsc{PoS} as a controlled testbed beyond the binary probing regime explored by prior work.

\section{Method and Materials}
\subsection{Dataset}\label{ssec:dataset}

We conducted our experiments on two datasets: a naturally occurring corpus and a small controlled dataset constructed for targeted evaluation.

\textbf{The GUM treebank.} For the naturally occurring data, we selected the UD English GUM treebank \citep{Zeldes2017}, annotated following the Universal Dependencies scheme.\footnote{\url{https://universaldependencies.org/}} We chose this treebank for its representativeness across diverse textual genres (academic, blog, legal, news, social, wiki, etc.), its medium size (14,353 sentences, 252,284 tokens), and its complete coverage of the 17 Universal PoS tags. The training split was used as the discovery set, while the test split was kept held out and used only to evaluate  whether discovered activations remain active on unseen tokens of the corresponding \textsc{PoS} categories.

\textbf{Controlled dataset.} To complement the naturally occurring data, we constructed a small controlled dataset of 180 lexical items to verify whether latents associated with specific \textsc{PoS} tags activate systematically in minimal, grammatically well-formed sentences. The dataset focuses primarily on nouns and verbs. For nouns, we selected 160 items spanning multiple semantic categories (e.g., mammals, birds, flowers, vehicles, etc.), evenly split between animate and inanimate referents. Each noun was instantiated in singular and plural form within neutral templates, including impersonal constructions such as \textit{There is a dog} and transitive constructions such as \textit{I see the dog} and \textit{I have a dog}. These templates vary determiner contexts, including indefinite articles, definite articles, and bare plurals. For verbs, we included 20 high-frequency verbs compatible with a minimal intransitive template (\textit{I + verb}, as in \textit{I walk}), balanced between 10 regular and 10 irregular past-tense forms, to limit the impact of morphological idiosyncrasies. All base sentences were augmented with two variants: one adding an adjacent adjective for noun sentences or adverb for verb sentences, and one appending punctuation to the augmented sentence. This allows us to assess whether additional \textsc{PoS} tokens introduce their own characteristic activations and whether these interact with those observed in the base sentence. All sentences were also instantiated in present and past tense to account for potential tense-driven effects.

\subsection{Processing Pipeline}\label{ssec:processing_pipeline}
In the following, we describe the processing pipeline to obtain SAE latent activations.

\paragraph{Model.} 
We experiment on \texttt{LLaMA-3-8B}. We employ the \texttt{EleutherAI/sae-llama-3-8b-32x} pre-trained model as our SAE. Both models are available on HuggingFace. The SAE model is trained and used via the Sparsify library.\footnote{\url{https://github.com/EleutherAI/sparsify}} The library is designed to follow the SAE implementation described in \newcite{gao2025scaling}. 

\paragraph{Token-level Activations Extraction.}
To extract token-level activations, we feed the raw sentence text to the model using its original subword tokenizer, and recover hidden state activations from the residual stream of layer 30 (last layer before the output) of the model. We encode such activations with the SAE to produce the sparse activation vectors for each subword. We obtain, for each subword, the fraction of SAE latents that fired on that subword, and their activation strength.  
Then, we align subword tokens and UD surface forms via character-span overlap: For each UD token with character span $[t_{\mathrm{start}},\, t_{\mathrm{end}})$, all subword tokens whose span $[s,\, e)$ satisfies $s < t_{\mathrm{end}}$ and $e > t_{\mathrm{start}}$ are identified as overlapping. The leftmost such subword token is designated the \textbf{anchor}, and its SAE activations are adopted as the representation of the corresponding UD token. Note that we chose the leftmost subword because it is the position at which the UD token's identity first becomes available to the model. Averaging over subwords may instead  dilute category-bearing activations with continuation-piece activations.
This yields, for each token, a dense SAE activation vector that is composed of all SAE latents that fired on the token and their activation strength.

\paragraph{Sparse Feature Matrix Construction.}
For the probing experiments, we construct a Sparse \textbf{SAE Feature Matrix} from dense token-level activations.
To do so, we consider the union of latents active across the entire treebank, which constitutes a subset of the full SAE latent space, spanning $4096 \times 32 = 131{,}072$ dimensions (i.e., LLM hidden size $\times$ SAE expansion factor). We therefore project all token representations into the common sparse vector space defined by the latents observed at least once in the Treebank. Concretely, we construct a feature matrix $\mathbf{X} \in \mathbb{R}^{N \times D}$, where $N$ is the number of tokens and $D =130{,}246$ is the number of attested latents, with entry $x_{i,j}$ set to the activation strength of latent $j$ on token $i$, and zero otherwise. The sparse matrix is the input to the probing classifiers.\footnote{\url{https://huggingface.co/datasets/colinglab/UD_English-GUM-Latents_Meta-Llama-3-8B_L30}}

\section{Experiments}\label{sec:experiments}

Our experiments are designed to assess not only whether \textsc{PoS} information is recoverable from \textsc{SAE} activations, but also how this information is organized in the latent space. In particular, we structure the analysis around the three research questions introduced in Section~\ref{sec:intro}. First, we test whether morpho-syntactic distinctions are explicitly available in the sparse activation space (\textbf{RQ1}). Second, we investigate the organization of \textsc{POS} categories in the latent space (\textbf{RQ2}). Third, we evaluate whether such organization is stable across splits and evaluation settings, and whether it supports general \textsc{PoS} classification (\textbf{RQ3}).

The experimental pipeline proceeds as follows. We first assess the linear recoverability of each \textsc{PoS} category with one-vs-rest probing classifiers (Section~\ref{sec:pos-probing}). We then localize \textsc{PoS}-relevant latent groups by combining feature salience, coverage, and compactness analyses (Section~\ref{sec:pos_localization}). Finally, we test the robustness of the selected groups on held-out data (Section ~\ref{sec:exp-heldout-data}).
This design separates recoverability, localization, and stability. Probing shows whether \textsc{PoS} information is present, while localization and validation assess how such information is organized in the latent space.

\subsection{Recoverability of \textsc{PoS} Information (RQ1)}
\label{sec:pos-probing}

We first test whether \textsc{PoS} distinctions are linearly recoverable from \textsc{SAE} activations. For each token in the GUM training split, we use the sparse \textsc{SAE} activation vector (cf. Section~\ref{ssec:processing_pipeline}) as input representation and the gold UD \textsc{PoS} tag as supervision.

We evaluate the one-vs-all setting using 5-fold cross-validation on the GUM Treebank Train split. For each \textsc{PoS} category, we train a binary classifier to distinguish tokens with that \textsc{PoS} from all other tokens. We train an L1-regularized logistic regression classifier (C = 0.1) using the liblinear solver, with balanced class weighting to account for label imbalance. The L1 penalty encourages sparse weight vectors, effectively performing feature selection and yielding interpretable models where most coefficients are driven to zero, given the high-dimensional nature of SAE latent spaces.

This allows us to assess the extent to which individual \textsc{PoS} distinctions are linearly recoverable from \textsc{SAE} activations.

\subsection{\textsc{PoS} Organization in Latent Space (RQ2)}
\label{sec:pos_localization}

We next ask how the \textsc{PoS} information recovered by the probes can be localized in the space of \textsc{SAE} latents. To this end, we use the one-vs-rest classifiers introduced in Section~\ref{sec:pos-probing} not only as predictive models, but also as feature-salience mechanisms.

\paragraph{Feature salience.}
For each \textsc{PoS} category, the corresponding logistic regression classifier assigns a coefficient \(\beta\) to each latent. Since \(\beta > 0\) indicates that the activation of a latent increases the probability of the positive class, we rank latents for each category according to their positive coefficients. We consider only latents with \(\beta > 0\), obtaining for each \textsc{PoS} tag a salience-ranked list of features that support the classification of that category. This analysis moves from recoverability to localization: rather than asking whether \textsc{PoS} information is present, we ask which sparse features contribute most to each distinction.

\paragraph{Coverage and compactness.}
We then quantify how compact each localized group is. For each \textsc{PoS} tag \(c\), let \(L^{(k)}_c\) denote the set of the top-\(k\) latents in its salience-ranked list, \(T_c\) the set of gold-label tokens tagged with \(c\), and \(a_\ell(t)\) the activation of latent \(\ell\) on token \(t\). We define the coverage of \(L^{(k)}_c\) as:

\[
\mathrm{Cov}_c(k) =
\frac{1}{|T_c|}
\sum_{t \in T_c}
\mathbf{1}
\left[
\exists \ell \in L^{(k)}_c : a_\ell(t) > 0
\right]
\]

\noindent{}Coverage measures the proportion of tokens of category \(c\) for which at least one of the top-\(k\) salient latents is active. We define the number of latents required to account for category \(c\) as the smallest \(k\) such that coverage reaches a target threshold \(\tau = 0.95\):

\[
k^{\star}_c = \min \{ k \in \mathbb{N} : \mathrm{Cov}_c(k) \geq \tau \}
\]

\noindent{}This gives an estimate of the effective size of the latent group associated with each \textsc{PoS} category. We use a per-class threshold rather than a global top-$k$ or coefficient threshold to avoid biasing the comparison due to high imbalance in i.) relevant latents for each \textsc{PoS} and ii.) coefficient profiles in open- vs closed-classes. Comparing \(k^{\star}_c\) across tags allows us to test whether different categories are represented with different degrees of compactness, for example whether open-class categories require broader latent groups than closed-class categories.

\paragraph{Compact-feature classification.}
Finally, we test whether the localized latent groups are sufficient for joint \textsc{PoS} prediction. Let \(C\) denote the set of \textsc{PoS} categories. We define the set of \textsc{PoS}-relevant latents as the union of the minimal coverage sets:
\[
L^{\star} =
\bigcup_{c \in C}
L^{(k^{\star}_c)}_c 
\]

\noindent{}We then train a multinomial logistic regression classifier using only \(L^{\star}\) as input features. This provides a stricter test of the localization procedure: if the selected latents capture systematic morpho-syntactic information, they should support multi-class \textsc{PoS} classification with limited degradation. A substantial drop with respect to the full \textsc{SAE} representation would instead suggest that relevant information remains distributed across additional latents.

\subsection{Validation on Held-Out Data (RQ3)}
\label{sec:exp-heldout-data}

We evaluate whether the latent groups identified in Section~\ref{sec:pos_localization} are stable beyond the data used to select them. The salience and coverage analyses are performed on the GUM training split, where \textsc{PoS} tags may correlate with lexical identity, frequency, position, or local syntactic patterns. We therefore test the selected groups in two complementary settings: the held-out GUM test split and the controlled dataset described in Section~\ref{ssec:dataset}.

To assess how specific each minimal latent group is to its target category in held out data we construct a cross-\textsc{PoS} activation matrix. For each pair of categories $(c, c'\in C)$, we compute the probability that at least one latent in $L^{(k^\star_{c'})}$ is active on tokens whose gold label is $c$:

$
M_{c,c'} =
\frac{1}{|T_c|}
\sum_{t \in T_c}
\mathbf{1}
\left[
\exists, \ell \in L^{(k^\star_{c'})} : a_\ell(t) > 0
\right]
$

This analysis assesses whether the POS-discriminative SAE latents are category-specific or shared across the various syntactic categories.

\subsection{Controls and Baselines}
We also provide a set of controls and baselines that address possible confounds and contextualise the SAE results.
First, we perform a control experiment to test whether \textbf{lexical identity} (i.e., the specific word form, like the preposition \emph{of}) can act as a confound for the probing experiments. To estimate how much of the original performance can be attributed to memorisation, we re-run the probe but assign each word type a random UPOS label. A probe relying solely on word identity would fit the control labels as well as the real ones.

Second, we provide several baselines for the probing experiment: i.) we use raw embeddings (layer 0) and raw activations from layer 30 as features for the probe instead of SAE activations; ii.) we provide a random latent subset baseline, where we re-run the compact-feature classification, but we keep a percentage of the $L\star$ (0, 25, and 50) and randomly choose the remaining latents.

\section{Results}
\label{sec:results}

\subsection{RQ1: Recoverability of \textsc{PoS} Information}
\label{sec:results-probing}

\begin{figure}[t]
    \centering
   \includegraphics[width=\columnwidth]{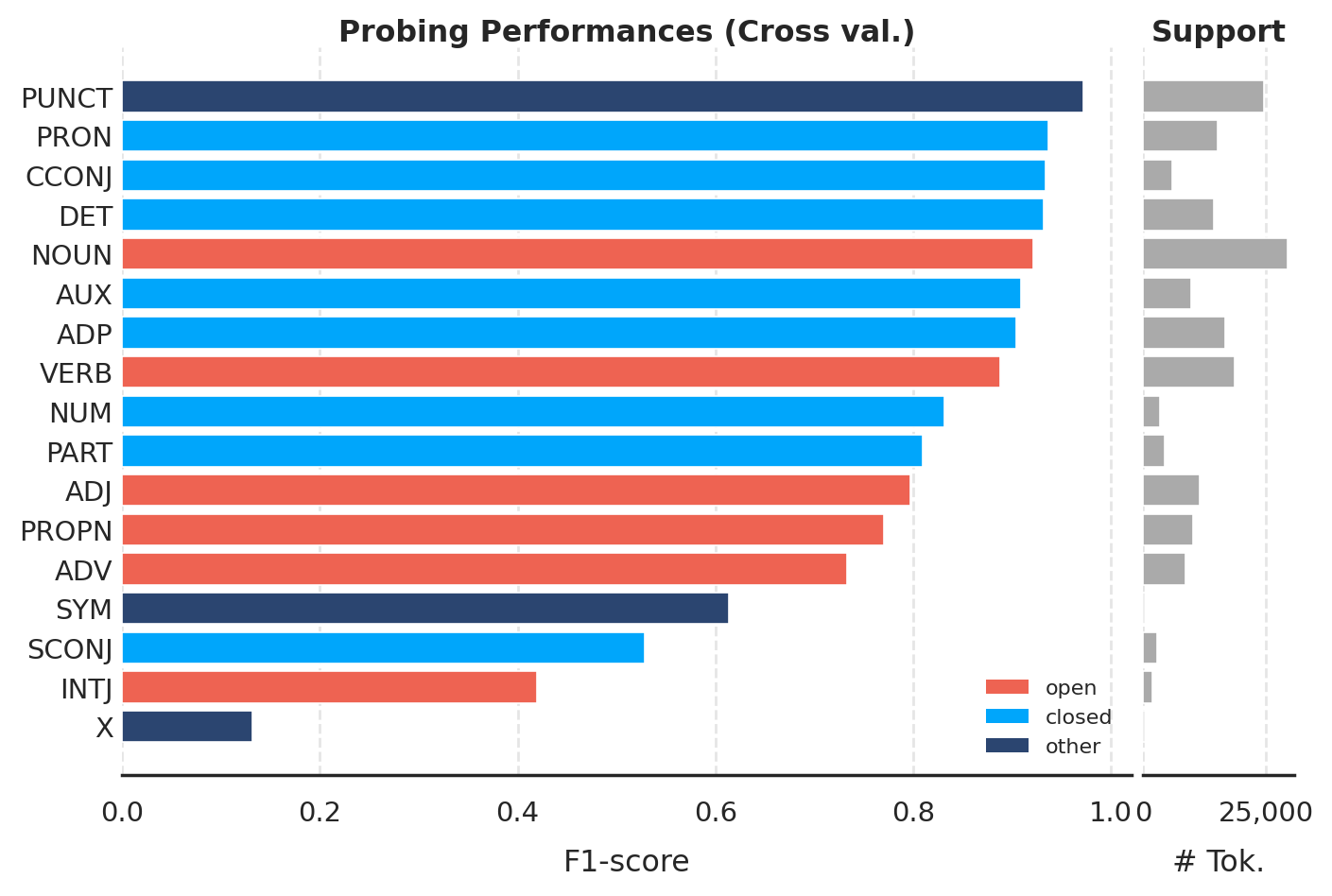}
\caption{One-vs-rest probing performance for each \textsc{PoS}. F1 scores are computed with 5-fold cross-validation on the GUM training split.}    \label{fig:probing1vs_rest}

\end{figure}

The one-vs-rest probing results show that \textsc{PoS} distinctions are consistently recoverable from \textsc{SAE} activations. As shown in Figure~\ref{fig:probing1vs_rest}, the binary classifiers achieve high F1 scores for most categories across the 5-fold cross-validation setting. This indicates that morpho-syntactic information is explicitly available in the sparse latent space.

Performance, however, is not uniform across tags. Closed-class categories and low-variability labels (e.g., punctuation), are easier to recover, while more lexically heterogeneous or less frequent categories show lower scores. Interestingly, nouns and verbs are the best performing open \textsc{PoS}. This suggests that recoverability is affected both by the linguistic nature of the category and by its support.

Overall, these results answer \textbf{RQ1} positively: \textbf{\textsc{SAE} activations contain information that is predictive of \textsc{PoS} categories.} At the same time, probing performance alone does not reveal how this information is organized within the SAE latent space.

\subsection{RQ2: \textsc{PoS}-related Latent Groups}
\label{sec:results-coverage}

\textbf{RQ2} asks whether the \textsc{PoS} information recovered by the probes is localized in restricted regions of the \textsc{SAE} latent space, and at what granularity. We report below the results of the three experiments introduced in Section~\ref{sec:pos_localization}.

\begin{figure}[t]
    \centering
    \includegraphics[width=\columnwidth,]{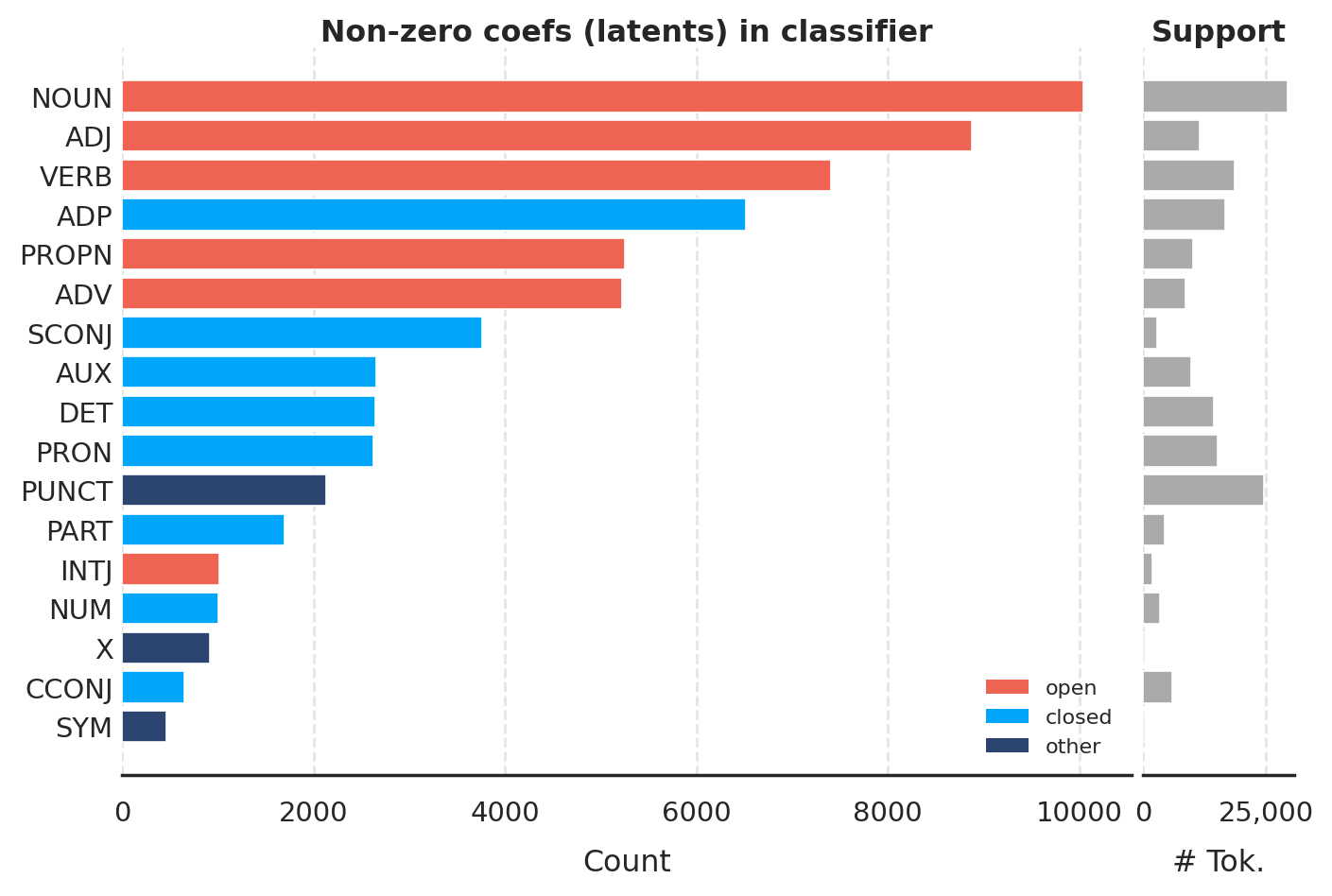}
    \caption{Non-zero coefficients for each one-vs-rest classifier; results are color coded by \textsc{PoS} class (Open, Closed, Other).}
    \label{fig:non-zero-coefs-1vsrest}
\end{figure}

\paragraph{Feature salience.}
The coefficient-based salience analysis shows that the binary probes do not rely uniformly on the full \textsc{SAE} latent space. For each \textsc{PoS} category, only a subset of latents receives positive weight, indicating that the classifier uses information concentrated in category-specific groups of sparse features. At the same time, these groups are not single-latent representations: \textbf{\textsc{PoS} distinctions are supported by multiple positively contributing latents, consistent with a distributed, but non-uniform, organization of morpho-syntactic information}.
Figure \ref{fig:non-zero-coefs-1vsrest} displays the number of non-zero coefficients for each one-vs-rest classifier. It emergers quite clearly that Open-class \textsc{PoS} have generally more non-zero latents, while Closed-class and Other-class have markedly less.\footnote{See Appendix \ref{app:results-coverage} for additional details.}

\begin{figure}[t]
    \centering
    \includegraphics[width=\columnwidth]{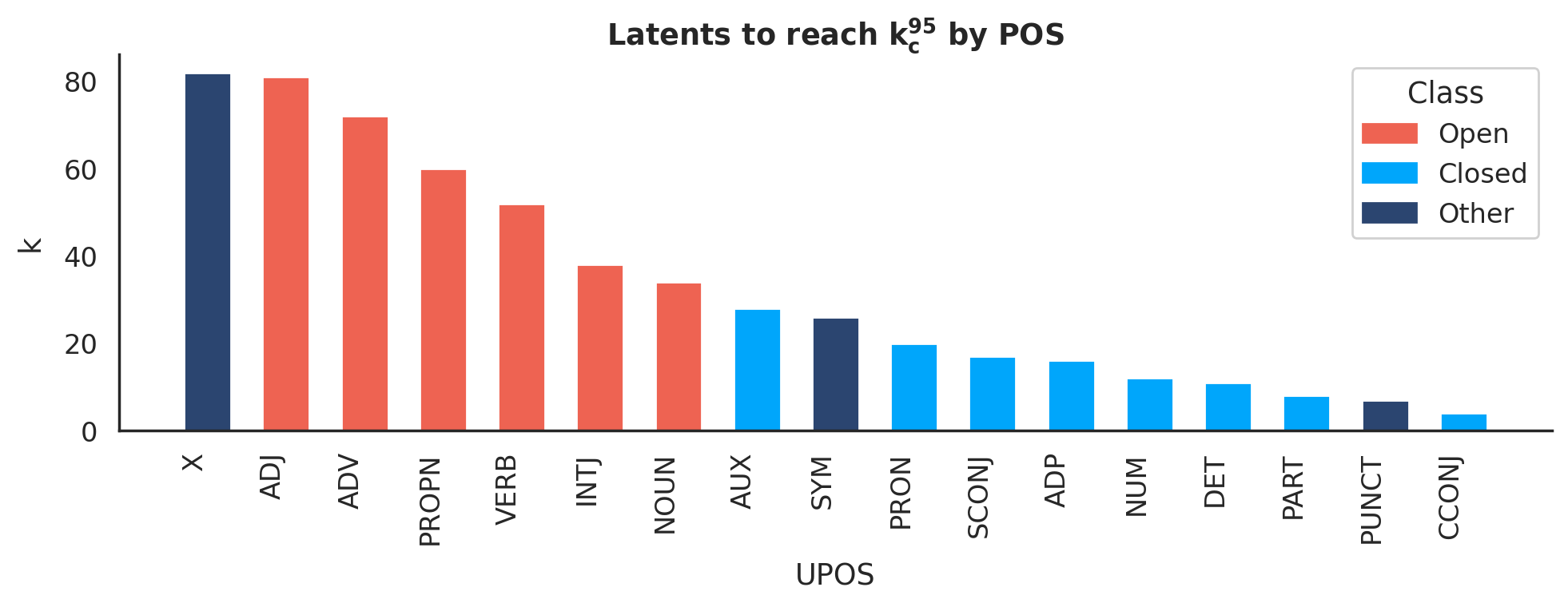}
    \caption{Number of salient latents required to reach 95\% coverage for each \textsc{PoS} category. For each tag \(c\), \(k_c^{95}\) denotes the smallest number of highest-coefficient latents needed to activate on at least 95\% of gold tokens of that category. Lower values indicate compact groups.}
    \label{fig:cardinality}

\end{figure}

\paragraph{Coverage and compactness.}
The coverage analysis further quantifies the effective size of these groups. As shown in Figure~\ref{fig:cardinality}, the cardinality \(k_c^{95}\) varies across \textsc{PoS} categories. Some tags reach the 95\% coverage threshold with a small number of latents, suggesting compact activation patterns. Others require broader latent groups, indicating that the corresponding distinction is more diffuse or depends on a wider set of lexical and contextual cues. This variability shows that localization is category-dependent and cannot be reduced to a one-latent-per-tag mapping. 

\begin{figure}[t]
    \centering
    \includegraphics[width=\columnwidth]{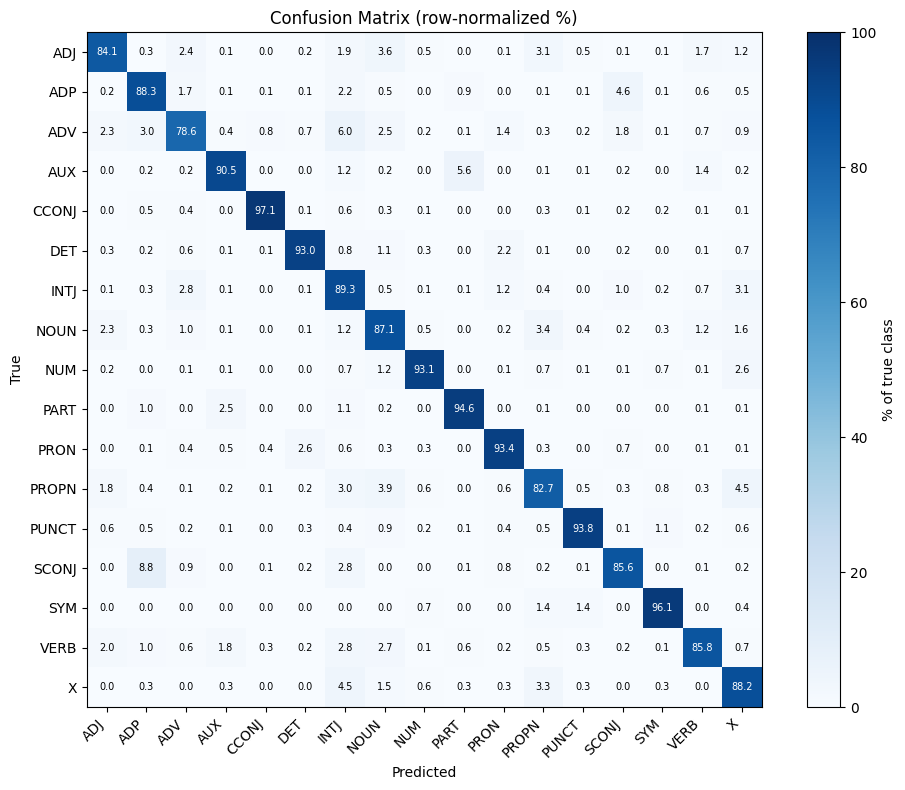}
    \caption{Row-normalized confusion matrix of the multi-class \textsc{PoS} classifier trained on the compact feature set \(L^\star\). Each cell reports the percentage of tokens of a gold \textsc{PoS} category predicted as each class.}
    \label{fig:probing_multiclass}
\end{figure}

\paragraph{Compact-feature classification.}
Finally, we test whether the selected latent groups are sufficient for multi-class \textsc{PoS} prediction. The compact feature set includes 498 latents, with 12\% of them being shared between 2+ \textsc{PoS}. The classifier trained on it achieves performance comparable to the classifier trained on the full \textsc{SAE} representation (Figure~\ref{fig:probing_multiclass}). The selected latents then preserve most of the information needed for multi-class \textsc{PoS} discrimination, and the salience and coverage analyses recover a compact but effective subset of the latent space.\footnote{See App. \ref{app:results-coverage} for a sensitivity analysis for $C$ and $\tau$.}

Overall, these results answer \textbf{RQ2} by showing that \textbf{\textsc{PoS} information is localized at the level of structured groups of latents}. These groups are compact for some categories and broader for others, but they are sufficient to support both category-wise coverage and multi-class classification.

\subsection{RQ3: Validation on held-out data}\label{sec:heldout-data}
We evaluate whether the latent groups identified in Section~\ref{sec:pos_localization} remain stable and systematic beyond the data used to select them, addressing \textbf{RQ3}.
We do not train another classifier, but rather test whether the latent groups identified in the discovery setting remain active on unseen instances of the corresponding \textsc{PoS} categories.
Recall that on held-out data we compute the probability that at least one latent in $L^{(k^\star_{c'})}$ is active on tokens whose gold label is $c$ For each pair of categories $c, c'\in C$.

\begin{table}[t]
\centering
\tiny
\begin{tabular}{lccc}
\toprule
\textbf{Template} & \textbf{P} & \textbf{R} & \textbf{F1} \\
\midrule
I see/saw [DET] [NOUN]                          & 0.207 & 0.984 & 0.341 \\
I see/saw [DET] [ADJ] [NOUN]                    & 0.215 & 0.967 & 0.351 \\
I see/saw [DET] [NOUN] [PUNCT]                  & 0.224 & 0.971 & 0.365 \\
I have/had ([DET]) [NOUN]                           & 0.186 & 0.977 & 0.313 \\
I have/had ([DET]) [ADJ] [NOUN]                     & 0.200 & 0.973 & 0.332 \\
I have/had ([DET]) [ADJ] [NOUN] [PUNCT]             & 0.212 & 0.976 & 0.348 \\
There is/was ([DET]) [NOUN]                    & 0.163 & 0.801 & 0.270 \\
There is/was ([DET]) [ADJ] [NOUN]              & 0.178 & 0.824 & 0.293 \\
There is/was ([DET]) [ADJ] [NOUN] [PUNCT]      & 0.193 & 0.846 & 0.314 \\
I [VERB]                                        & 0.162 & 0.975 & 0.278 \\
I [VERB] [ADV]                                  & 0.175 & 0.970 & 0.297 \\
I [VERB] [ADV] [PUNCT]                          & 0.191 & 0.975 & 0.319 \\
\midrule
\textbf{Average}                                & \textbf{0.192} & \textbf{0.937} & \textbf{0.318} \\
\bottomrule
\end{tabular}
\caption{Pseudo-multilabel classification results on the controlled dataset. }
\label{tab:templates}

\end{table}

On the controlled dataset, which includes only a subset of \textsc{PoS} categories, we evaluate the selected latent groups as a pseudo-multilabel classification task. For each token, we encode its gold \textsc{PoS} as a binary vector over the considered categories $C$. Predictions are obtained by applying the indicator function defined in Section~\ref{sec:heldout-data} to each category $c' \in C$. Table~\ref{tab:templates} reports micro-averaged precision, recall, and F1 for each template. The results show limited precision but very high recall in general, with some template-dependent variation.  We also note the presence of a set of latents that we associate with a ``first-token'' concept and that confound the results. We verified that by including a prefix with a different \textsc{PoS} to each sentence, all these activations shift onto this new \textsc{PoS},\footnote{We report an example of this in Appendix \ref{app:first-tok-act}}
indicating that latent groups are present in controlled data.

\begin{figure}[t]
    \centering
   \includegraphics[width=\columnwidth]{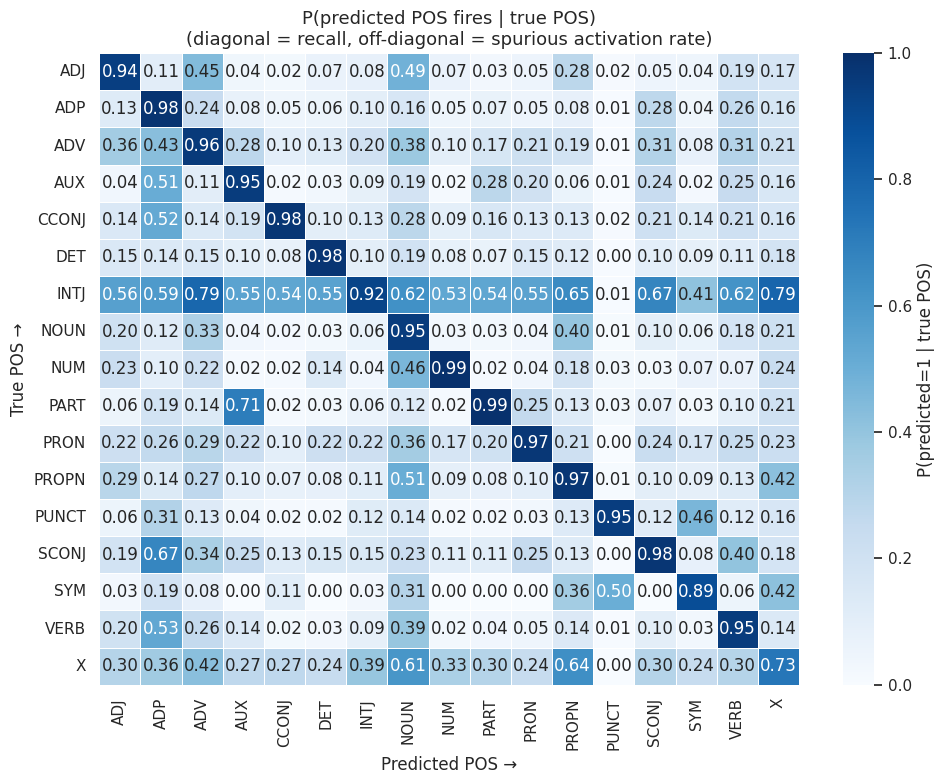}
    \caption{Co-activations of selected latent groups on the held-out treebank test set. Diagonal values correspond to recall for each \textsc{PoS}; off-diagonal values correspond to false positive rates on other \textsc{PoS} categories.}
    \label{fig:heldout_treebank}

\end{figure}

For more generalizability over the whole \textsc{PoS} set, we use the held-out treebank test set. Figure~\ref{fig:heldout_treebank} reports a heatmap with  the co-activation patterns. By construction, the diagonal entries $M_{c,c}$ represents the per-category recall. The off-diagonal entries $M_{c,c'}$ ($c \neq c'$) measure the rate at which latents selected as salient for $c'$ nevertheless fire on tokens of a different category $c$, i.e., a spurious co-activation rate. Low off-diagonal values indicate that the minimal latent groups are disjoint and category-specific; high values suggest overlapping representations across \textsc{PoS} categories. Results highlight two main aspects: First, the recall is consistently high, with most values $\geq 0.95$, indicating that the latents identified as salient for a given \textsc{PoS} remain active on unseen tokens drawn from the same underlying distribution; second, we observe a relatively high variability in off-diagonal scores, indicating overlaps across categories. We further validate the latter observation by computing a \textit{Distinctiveness} ($D$) score for each \textsc{PoS}, as $\mathrm{D}(c) = \frac{M_{c,c}}{\sum_{c' \in C} M_{c,c'}}$. Intuitively, $\mathrm{D}(c) = 1$ indicates that the latents in $L^{(k^\star_c)}$
fire exclusively on $c$ tokens; a value of $1/|C|$ (0.06 in our case) corresponds to
the chance baseline under uniform co-activation. We observe a $\mathrm{D}$ $\mu=0.27$ ($\sigma=0.07$).\footnote{See Appendix \ref{app:distinctiveness} for the per-$c$ results.}

Overall, the results on both held-out datasets answer \textbf{RQ3} affirmatively for stability, while qualifying the systematicity claim: the identified latent groups remain consistently active on unseen tokens of their target category, but are only partially category-specific. More generally, this suggests that, \textbf{despite the transition to a sparse representation through the \textsc{SAE}, a one-to-one correspondence between linguistic categories and groups of latents holds only to a limited extent}.

\begin{figure}[t]     
\centering \includegraphics[width=0.95\columnwidth]{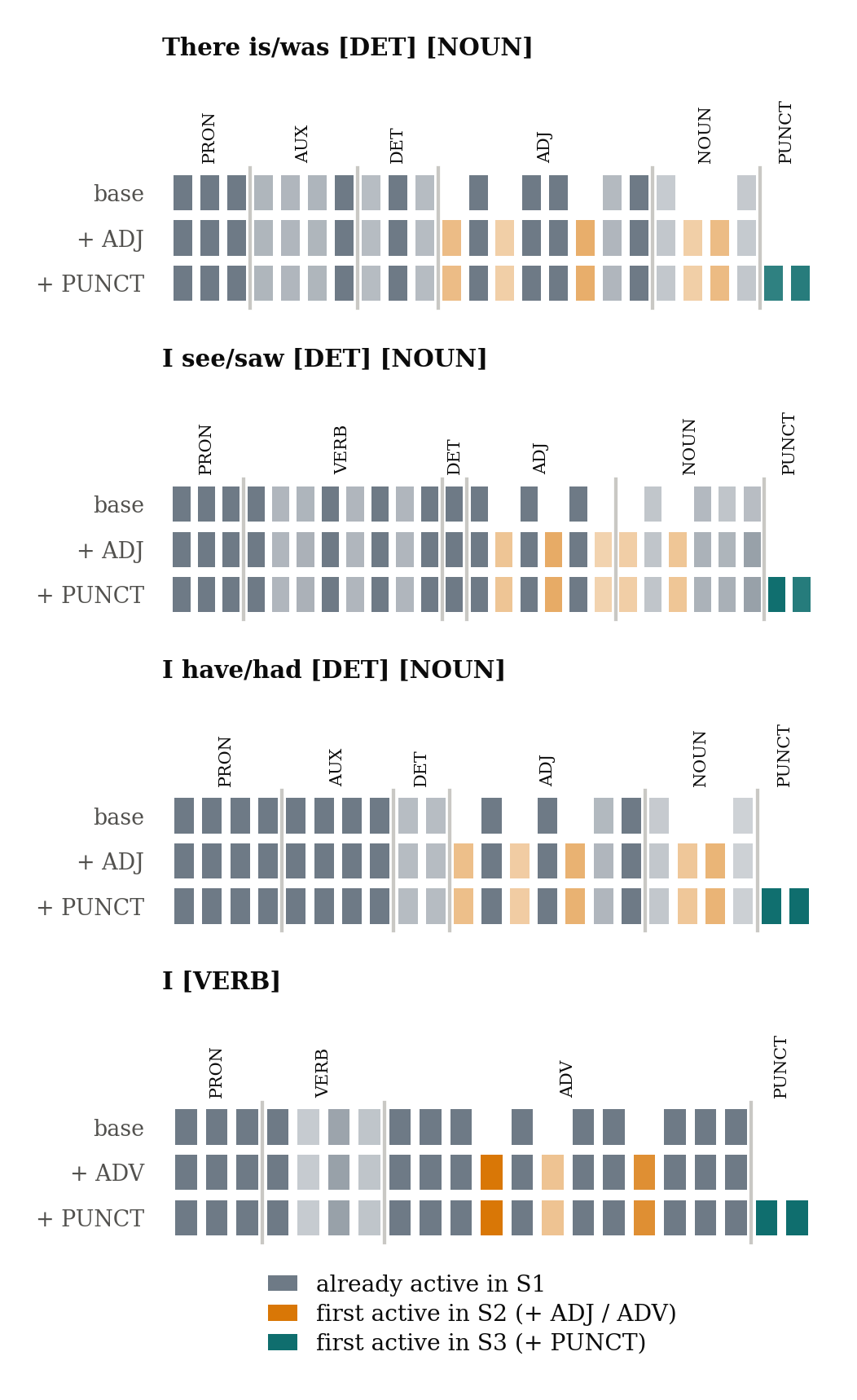}
    \caption{Active L$\star$ latents per relevant \textsc{PoS} in each template variant. Each square is one latent that is active on at least 25\% of template's examples. Color saturation indicates activation percentage (darker $=$ more active).}
    \label{fig:sent_col_comparison_there_is_are}

\end{figure}

\subsection{Controls and Baselines}

The control probe with random labels obtains 0.54 Accuracy/0.42 Macro F1, against 0.88/0.97 respectively for the real probe. The confound is not negligible, but the 34–37 point performance gap still supports the conclusion that POS recoverability is not simply reducible to lexical memorisation, and that \textbf{lexical identity plays a minor role}. This is consistent with evidence provided in Sec. \ref{sec:ling_analysis}.

The \textbf{SAE-based probe has comparable performances with both the raw residual stream probes}, both at the embedding layer and at layer 30 (Table \ref{tab:probes-comparison}) using $\sim$8 times less features. However, note that we do not claim superiority of SAEs as a probing tool. Rather, we claim and show that (i) POS information is recoverable from subsets of SAE latents and that (ii) the SAE's contribution is decomposition and localisation, which the dense probe cannot provide as easily.

\begin{table}[]
\centering
\scriptsize
\begin{tabular}{lcc}
\hline
\textbf{Probe}      & \textbf{Accuracy} & \textbf{Macro F1} \\ \hline
SAE        & 0.88     & 0.78     \\
Layer 30   & 0.92     & 0.84     \\
Layer 0 (Embed.) & 0.88     & 0.78     \\ \hline
\end{tabular}
\caption{Performances using different probes.}
\label{tab:probes-comparison}
\end{table}

Finally, we see that \textbf{$L^\star$ features are \textsc{PoS} relevant.} In fact, randomly replacing features from the $L^\star$ set drastically reduces performances. Table \ref{tab:L_star_overlap_eval} shows the results at 0, 25, 50 and 100\% overlap with $L^\star$. This further demonstrates that $L^\star$ latents are \textsc{PoS} relevant.

\begin{table}[t]
\centering
\scriptsize
\begin{tabular}{llcc}
\toprule
\textbf{Overlap (\%)} & \textbf{Evaluation} & \textbf{Accuracy} & \textbf{Macro-F1} \\
\midrule
\multirow{2}{*}{0}
 & Cross-validation & 0.24 & 0.16 \\
 & Train/test       & 0.23 & 0.16 \\
\midrule
\multirow{2}{*}{25}
 & Cross-validation & 0.49 & 0.41 \\
 & Train/test       & 0.49 & 0.41 \\
\midrule
\multirow{2}{*}{50}
 & Cross-validation & 0.69 & 0.60 \\
 & Train/test       & 0.69 & 0.60 \\
\midrule
\multirow{2}{*}{100 (original)}
 & Cross-validation & 0.87 & 0.76 \\
 & Train/test       & 0.89 & 0.81 \\
\bottomrule
\end{tabular}
\caption{Accuracy and Macro-F1 across overlap levels, comparing cross-validation and train/test evaluation.}
\label{tab:L_star_overlap_eval}
\end{table}

\subsection{Linguistic analysis of SAE activations}\label{sec:ling_analysis}

The results confirm a systematic association between \textsc{SAE} activations and \textsc{PoS} information. Overall, four main patterns emerge.

\textbf{Closed classes are more stable and compact.}
On the training set (Figure \ref{fig:probing1vs_rest}) and on the held-out treebank test set (Figure \ref{fig:heldout_treebank}), Closed-classes consistently achieve higher Recall and F1-score. Note that these categories also require fewer latents in the coverage analysis (Figure~\ref{fig:cardinality} and Appendix~\ref{app:results-coverage}, Figure~\ref{fig:non-zero-coefs-1vsrest}). By contrast, \textsc{x} and \textsc{sym} exhibit the highest error rates in both experiments, likely due to their low support in the training data (Figure \ref{fig:probing1vs_rest}).

The compactness gradient we observe co-varies with the size and formal variability of each category's type inventory. A category with a handful of invariant word forms can be covered by a small latent group with no category-level abstraction being involved, since a group of form-specific latents suffices. The controlled data support this reading: DET collapses to a single activation value where the determiner is invariably the (Figure \ref{fig:sent_col_comparison_I_see_saw}), and acquires structure only where the a/an versus bare-plural alternation introduces formal variation (Figure \ref{fig:sent_col_comparison_there_is_are}). The compactness ordering should therefore be read primarily as a gradient in lexical variability rather than as direct evidence of graded abstraction.

\textbf{Open classes show broader and less selective activation patterns.}
Spurious co-activations are more frequent for open \textsc{PoS} classes, where the same lemma or morphologically related forms can serve different functions depending on context. For instance, \textsc{adj} tokens are mainly confused with \textsc{noun}, \textsc{adv}, and \textsc{propn}, reflecting attributive noun uses, adjective--adverb overlap, and nominal modification patterns. Similarly, \textsc{adv} shows diffuse co-activation with \textsc{adp}, \textsc{noun}, \textsc{adj}, \textsc{sconj}, and \textsc{verb}, suggesting that some adverb-associated latents capture positional or contextual cues rather than adverbial function alone. \textsc{propn} and \textsc{noun} also co-activate, consistent with their shared nominal distribution, while \textsc{verb} shows overlap with \textsc{noun} in homograph pairs e.g. \textit{to drink} / \textit{the drink}.

\textbf{Some off-diagonal patterns reflect annotation and lexical overlap, as well as syntagmatic properties.}
The strongest non-target activations are not random. \textsc{intj} shows high spurious activation rates across several categories, plausibly due to annotation conventions that assign heterogeneous forms such as \textit{like}, \textit{well}, or \textit{God} to \textsc{intj} in pragmatic contexts. Similarly, \textsc{sconj} co-activates with \textsc{adp} and \textsc{verb}, reflecting lexical overlap between subordinating conjunctions and prepositions in English (e.g., \textit{by}, \textit{after}, \textit{since}) and broader positional regularities. These patterns suggest that the selected latents do not encode purely abstract \textsc{PoS} function, but also respond to surface form, lemma sharing and orthographic cues. Moreover, the \textsc{Pos} emerging out of the latent space are also defined in terms of their syntactic contexts. For instance, the \textsc{adj} latents strongly co-activate with \textsc{noun} reflecting the nature of adjectives as nominal modifiers.

\textbf{Controlled examples confirm additive and category-sensitive activations.}
Turning to the controlled dataset, Figure~\ref{fig:sent_col_comparison_there_is_are} confirms that, as new tokens are introduced into the sentence, their associated latents activate consistently within the latent groups characteristic of their PoS,
suggesting that \textbf{PoS-specific latent activations are largely additive across tokens.} For example, adding \textit{loyal} to  \textit{There is a dog} triggers the activation of more latents associated with \textsc{adj}. 
We observe that some latents of related \textsc{PoS}es are already present even without the corresponding words (e.g., \textsc{adj} for \textsc{noun} and \textsc{adv} for \textsc{verb}), but the number of active ones corresponding to that category consistently grows when the word is included in the template. 

These findings show that \textsc{PoS}-related latent groups are stable and systematic, but not category-exclusive. Closed classes tend to yield compact and selective representations, whereas open classes involve broader latent groups that also capture lexical, morphological, and contextual regularities.


\section{Conclusion}
We used \textsc{PoS} categories as a controlled testbed to study how morpho-syntactic information is organized in \textsc{SAE}s. We show that \textsc{PoS} distinctions are consistently recoverable from sparse activations, and that each category is supported by a compact group of sparse features, whose size varies with the linguistic nature of the category. For Open-class \textsc{PoS}, these groups are more diffuse, with a larger number of active latents, while Closed-class ones have fewer active latents. A small union of these category-specific latents preserves strong multi-class classification performance, and the selected groups remain stable on held-out treebank data and largely additive on controlled examples.

Our results reveal that \textsc{PoS} are internally represented in LLMs as emerging sets of localizable but distributed features in latent SAE space. Moreover, analyses suggest that interpretability claims at the latent level should be evaluated against theoretically grounded category inventories rather than top-activating examples alone. At the same time, cross-category co-activations show that the identified latents partly track lexical, positional, and annotation-driven regularities, motivating future work on richer linguistic levels and typologically diverse languages.

\section*{Limitations}
\label{sec:limitations}

Our study focuses on \textsc{PoS} categories as a controlled morpho-syntactic test case. While this choice allows us to rely on exhaustive and independently annotated labels, \textsc{PoS} tags capture only one level of linguistic abstraction. Future work should extend the analysis to finer-grained morphological features, dependency relations, semantic roles, and discourse-level phenomena, where latent organization may be recoverable in different ways.

We also analyze a single base language model, \textsc{LLaMA}-3-8B, and one publicly available \textsc{SAE}. The observed patterns may depend on the underlying model, the layer from which activations are extracted, the \textsc{SAE} training procedure, and the sparsity regime. Comparing multiple models, layers, and \textsc{SAE} variants would be necessary to assess how general these findings are.

Our localization procedure relies on linear probing coefficients as a salience signal. Although the use of \(\ell_1\)-regularization encourages sparse and interpretable solutions, probe coefficients should not be interpreted as direct causal evidence. The identified latents are predictive of \textsc{PoS} categories, but further causal interventions would be needed to establish whether they are used by the model for morpho-syntactic processing.

Finally, our controlled dataset is intentionally small and targets a restricted set of constructions, mainly involving nouns and verbs. It is useful for validating whether selected latent groups remain active in simple and independently constructed contexts, but it does not cover the full syntactic and lexical variability of English. Broader controlled datasets would allow a more systematic evaluation of how lexical ambiguity, word order, morphology, and sentence complexity affect \textsc{PoS}-related latent activations.

\section*{Acknowledgments}

This work has been supported by i.) the PNRR MUR project \href{https://fondazione-fair.it/}{PE0000013-FAIR} (Spoke 1), funded by the European Commission under the NextGeneration EU programme; ii.) the EU EIC project \href{https://eic-emerge.eu}{EMERGE} (Grant No. 101070918); and iii.) The PNRR MUR project FAIR TT\_02 ``Innovare
la sorveglianza automatizzata delle infezioni del sito chirurgico tramite modelli di elaborazione del linguaggio naturale''.

\bibliography{custom}

\newpage
\appendix

\section*{Appendix}

\section{Further Details on Implementation}
Here we provide some further details on several of the implementation choices, including the rationale for selecting the target layer and experimental setup.

\subsection{Layer Selection}
We extract hidden state activations of the LLM from the residual stream of the last layer before the output. The residual stream hookpoint is set after the MLP layer. This means that the activations we take are very close to the output. 

\begin{figure*}[t]
    \centering
    \includegraphics[width=\linewidth]{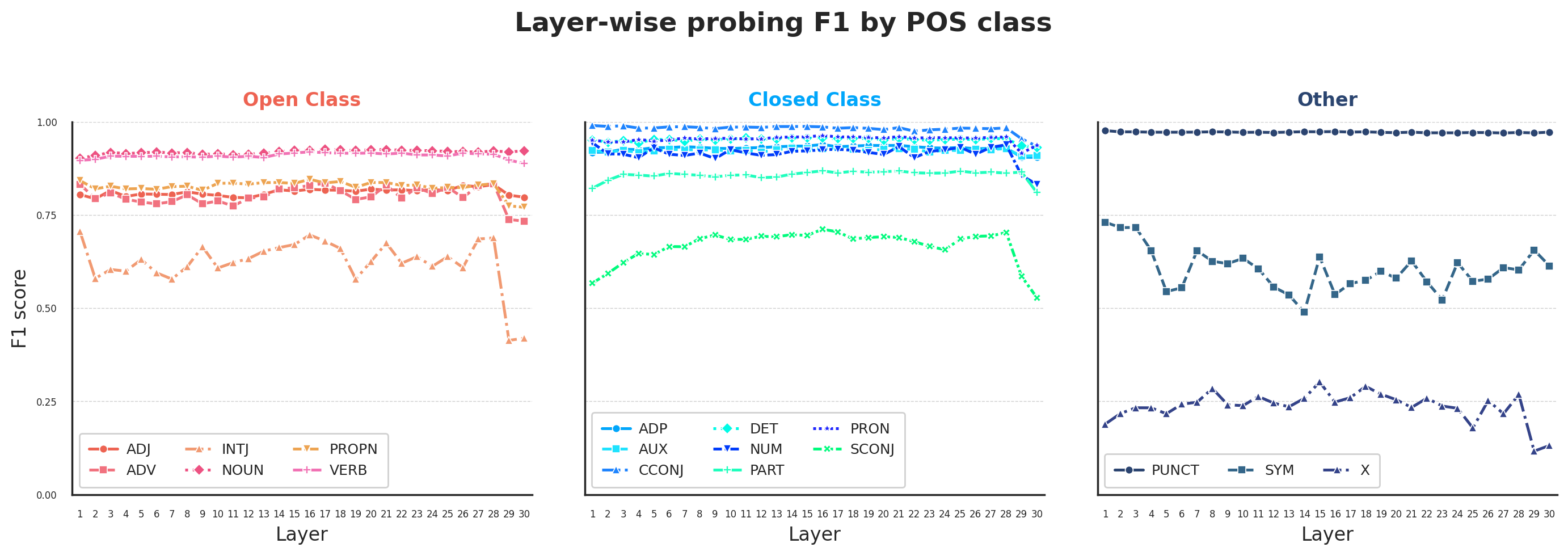}
    \caption{Per-\textsc{PoS} one-vs-rest classifier performances across layers.}
    \label{fig:layer-wise-performances}
\end{figure*}

We acknowledge that the choice of the layer might drastically affect the results. Thus, to verify the stability of our findings, we replicate some of the experiments across a sample of layers. As for  RQ1 (Recoverability of \textsc{PoS} information, see Section \ref{sec:pos-probing}), we extract hidden representations from the residual stream after the MLP of layer $x$, and encode it with the corresponding SAE. 
We report in Figure \ref{fig:layer-wise-performances} the F1-Scores of the binary probing classifiers across all layers. The results show that performances are generally stable across layers, with minor variations. Other class \textsc{PoS}es are the ones showing higher variability, except \textsc{Punct}. We also note a systematic slight dip in performances in the last few layers across all \textsc{PoS} classes. This may be attributable to the vicinity to the unembed layer.

As for RQ2 (Organization in Latent Space, see Section \ref{sec:pos_localization} and RQ3 (Validation on held-out data, see Section \ref{sec:exp-heldout-data}), we report result for Salience (latent activations count), Coverage (, Compact-feature classification, and distinctiveness for Layers 2 and 15 (beginning and middle of the model stack).
First, \textbf{the number of non zero activations per POS increases modestly across layers for most POS} (e.g., NOUN: 9024$\rightarrow$9935$\rightarrow$10043; ADP: 4955$\rightarrow$5742$\rightarrow$6511), \textbf{but stays within the same order of magnitude}: open-class POS in the low-thousands-to-10k range, closed-class in the hundreds-to-low-thousands, Other in the hundreds. Pooled mean rises only ~19\% from layer 2 to layer 30 (3105$\rightarrow$3697). This reflects gradual densification with depth, not a qualitative reorganization.
Second, \textbf{$L^*$ remains stable stable across depth}: 530 (L2) $\rightarrow$ 501 (L15) $\rightarrow$ 498 (L30). Coverage efficiency ($L^\star$ over the sum of $k^{\star}_c$) improving slightly (84\%$\rightarrow$81\%$\rightarrow$87\%). Per-category values do redistribute a bit. For example, ADP falls (71$\rightarrow$35$\rightarrow$16), SCONJ has a non monotonic trend (7$\rightarrow$51$\rightarrow$17), NOUN rises (12$\rightarrow$38$\rightarrow$34), PROPN rises (22$\rightarrow$47$\rightarrow$60). However, all values remain in the same single-to-double-digit-tens range at every layer, with no category jumping an order of magnitude.
Third, \textbf{compact-feature classification Macro F1 increases only slightly} (0.77$\rightarrow$0.79$\rightarrow$0.81), and accuracy stays mostly flat (0.87$\rightarrow$0.88$\rightarrow$0.87). This may be an indication that later layers encode more linearly separable POS distinctions for minority/harder classes. Finally, \textbf{distinctiveness remains stable across layers}. We report mean and standard deviation: 0.15$\pm$0.03 (L2) $\rightarrow$ 0.32$\pm$0.09 (L15) $\rightarrow$ 0.27 $\pm$ 0.07 (L30).

These results suggest that the layer choice do not drastically affect the degree to which \textsc{PoS}es are encoded into the SAE latents, and that the findings of the paper should hold across the whole model. Results on additional layers at the beginning and middle of the model stack show slight variation, but are also a clear indication that the conclusions from the paper hold also for layers other than the one analyzed. Individual POS categories reshuffle which latents they rely on across depth, but the total representational budget and per-category magnitudes remain stable. This supports our claim of layer-consistent POS structure with depth-wise redistribution rather than qualitative change.

\subsection{Experimental setup}\label{app:setup}
All experiments were conducted using a GPU node equipped with 8 A100 80GB GPUs. Given the model sizes, only one GPU was sufficient to extract hidden representations from a layer and encode it with its corresponding SAE. The process takes roughly 0.16 GPU hours per layer. 
The probing classifiers were implemented using SciKit-Learn. The library does not leverage GPUs, but allows parallelization across CPU cores. A single 5-fold cross-validation training/test run on the GUM Treebank training set requires roughly 4 hours using all CPU cores available.

\subsection{Artifacts and Intended Use}
\label{app:artifacts}

We use three existing artifacts, employed consistently with their intended use and license.

\paragraph{LLaMA-3-8B} \citep{dubey-2024-llama3} is released by Meta under the Llama~3 Community License, which permits research and academic use. We use the model exclusively for interpretability analysis, extracting hidden representations without fine-tuning or redistribution of model weights.

\paragraph{EleutherAI SAE} The checkpoint we use from \texttt{EleutherAI/sae-llama-3-8b-32x} is a pretrained Sparse Autoencoder released on HuggingFace by EleutherAI for interpretability research on LLaMA-3-8B,  Our use directly aligns with its intended purpose.

\paragraph{The GUM treebank} \citep{Zeldes2017} is distributed under Creative Commons licenses (primarily CC~BY~4.0, with some subcorpora under more restrictive terms) for research and educational use in computational linguistics. We use the treebank's text and Universal Dependencies annotations for probing and evaluation, which is consistent with its intended use.

\paragraph{Artifacts produced.} We release\footnote{\url{https://github.com/colinglab/pos-sae-latents}, \\
\url{https://huggingface.co/datasets/colinglab/UD_English-GUM-Latents_Meta-Llama-3-8B_L30}} the token-level SAE activations aligned to UD POS tags, the controlled evaluation dataset (180 items), and the code for the probing and salience pipeline. These artifacts are released for research use only, consistent with the access conditions of the underlying resources. To respect the per-source licensing of GUM, activation files are keyed to token indices in the original GUM release rather than redistributing the source text.

\section{Additional Results}\label{app:additional-results}

Here we present additional results from our experiments across our three RQs.

\subsection{RQ1: Recoverability of \textsc{PoS} information}\label{app:rq1}

\begin{figure}[t]
    \centering
    \includegraphics[width=\columnwidth,]{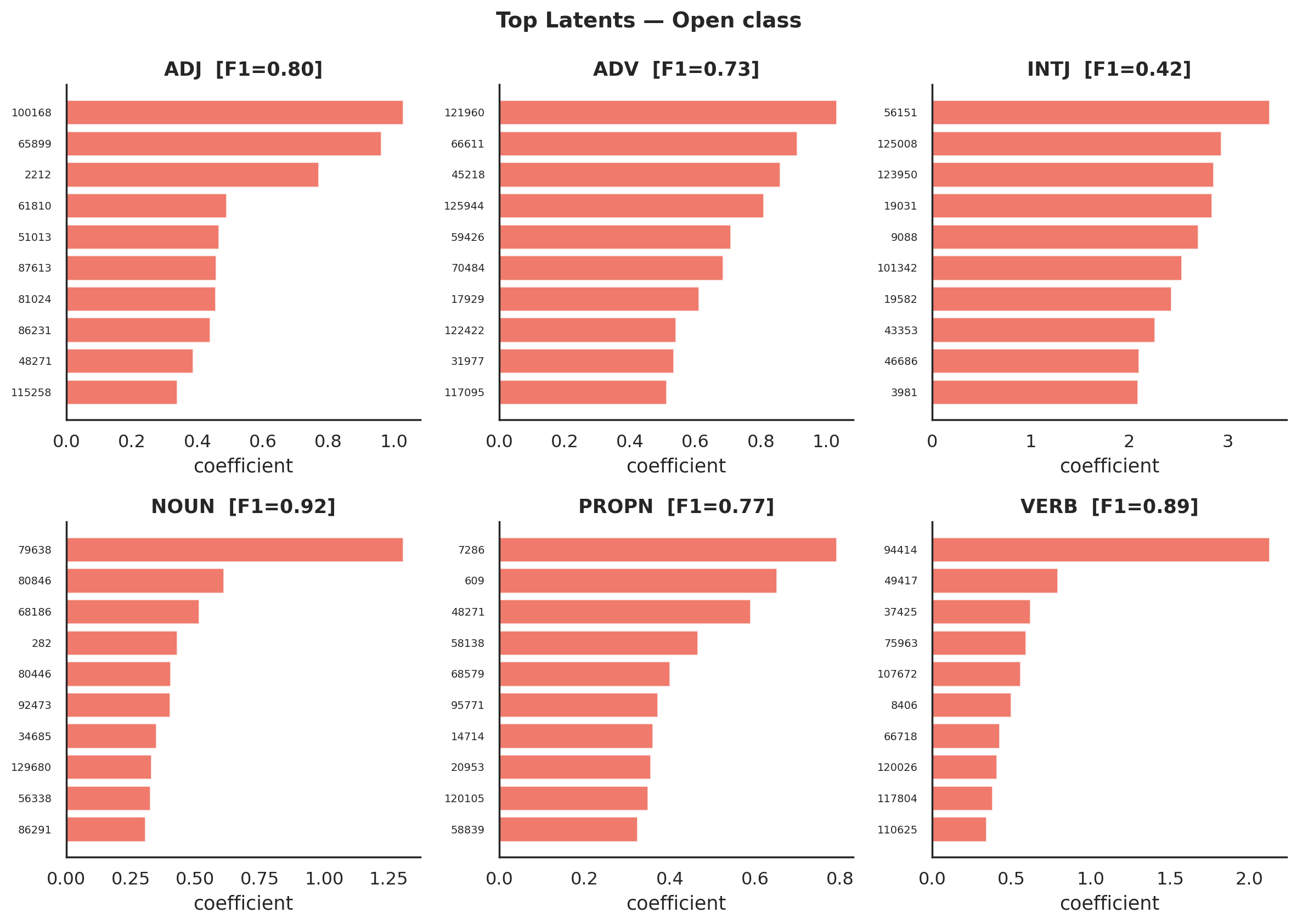}
    \caption{Per-\textsc{PoS} one-vs-rest classifier top coefficients. \textsc{Open} class \textsc{PoS}es.}
    \label{fig:top-latents-open}
\end{figure}

\begin{figure}[t]
    \centering
    \includegraphics[width=\columnwidth,]{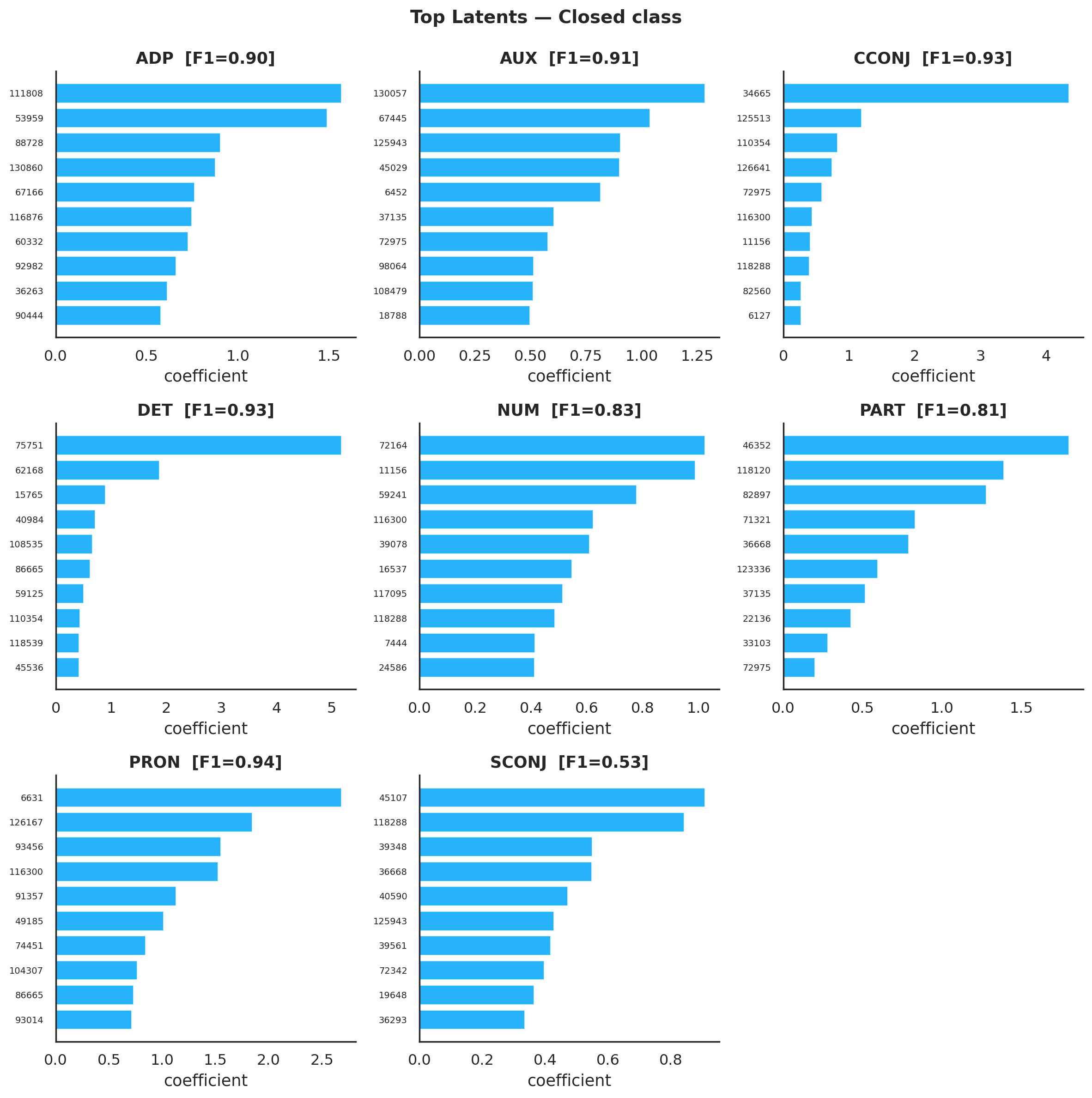}
    \caption{Per-\textsc{PoS} one-vs-rest classifier top coefficients. \textsc{Closed} class \textsc{PoS}es.}
    \label{fig:top-latents-closed}
\end{figure}

\begin{figure}[htbp]
    \centering
    \includegraphics[width=\columnwidth,]{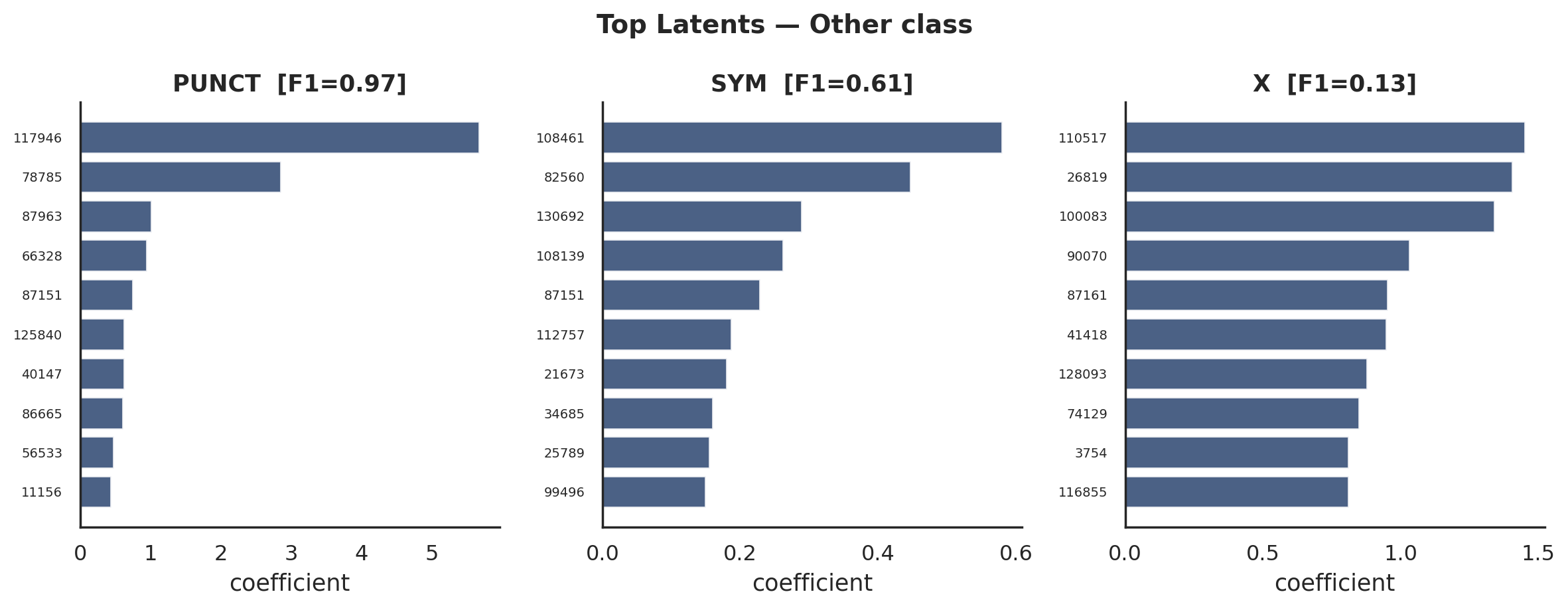}
    \caption{Per-\textsc{PoS} one-vs-rest classifier top coefficients. \textsc{Other} class \textsc{PoS}es.}
    \label{fig:top-latents-other}
\end{figure}

Figures \ref{fig:top-latents-open}, \ref{fig:top-latents-closed}, and \ref{fig:top-latents-other} show, for each \textsc{PoS} class, the top 10 latents ranked by coefficient in the classifier. 

Closed-class categories show the most peaked coefficient distributions, with a single latent dominating in \textsc{det} (75751, $\approx\!5.5$), \textsc{CCONJ} (34665, $\approx\!4.3$) and \textsc{PRON} (6631, $\approx\!2.7$). Open-class categories are flatter: only \textsc{VERB} shows a clearly dominant feature (94414, $\approx\!2.2$); \textsc{ADJ} (0.80), \textsc{PROPN} (0.77) and \textsc{ADV} (0.73) have no single salient latent. \textsc{PUNCT} is the best-classified category overall, driven by latent 117946 ($\approx\!5.5$). Moreover, Several latents recur across closed-class categories (e.g.,~72975 in \textsc{AUX}/\textsc{CCONJ}/\textsc{PART}; 116300 in \textsc{CCONJ}/\textsc{NUM}/\textsc{PRON}; 86665 in \textsc{DET}/\textsc{PRON}/\textsc{PUNCT}), which may indicate poly-functional latents.

\subsection{RQ2: \textsc{PoS}-related Latent Groups}
\label{app:results-coverage}

Here, we present results related to RQ2.

\paragraph{Feature Salience.} As for the feature salience, Table \ref{tab:nonzero-pos-latents} shows the number of non-zero latents for each one-vs-rest classifier. See also Figure \ref{fig:non-zero-coefs-1vsrest} in the main paper. From the Table and Figure, it emerges quite clearly that Open-class \textsc{PoS} have generally more non-zero latents, while Closed-class and Other-class have markedly less. The two main exceptions are \textsc{INTJ} for the Open-class, which has very few, and \textsc{ADP} for the Closed-class, which is more akin to open ones. For \textsc{ADP}, this may be attributable to the fact that words that function as adpositions may also be used to mark adverbial clauses. As for \textsc{INTJ}, they typically express an emotional reaction and are not syntactically related to other accompanying expressions; moreover, their support in the GUM treebank is very low. Both factors could play a role in poor classification performances (See Section \ref{sec:results-coverage} and below) and limited number of latents active as features during classification. Table \ref{tab:feat_salience} reports mean and standard deviation counts for non-zero coefficients. Again, we observe that Other class have the least number of non-zero coefficients on average, and Open class has the highest number and highest variability.

\begin{table}[htbp]
\centering
\scriptsize
\begin{tabular}{llcc}
\toprule
\textbf{POS} & \textbf{POS Type} & \textbf{Non-zero Latents} & \textbf{Frac. of Latent Space} \\
\midrule
NOUN  & open   & 10043 & 0.0766 \\
VERB  & open   & 7404  & 0.0565 \\
ADJ   & open   & 8877  & 0.0677 \\
PROPN & open   & 5252  & 0.0401 \\
ADV   & open   & 5218  & 0.0398 \\
INTJ  & open   & 1010  & 0.0077 \\
PRON  & closed & 2623  & 0.0200 \\
CCONJ & closed & 645   & 0.0049 \\
DET   & closed & 2647  & 0.0202 \\
AUX   & closed & 2651  & 0.0202 \\
ADP   & closed & 6511  & 0.0497 \\
NUM   & closed & 1001  & 0.0076 \\
PART  & closed & 1694  & 0.0129 \\
SCONJ & closed & 3762  & 0.0287 \\
PUNCT & other  & 2129  & 0.0162 \\
SYM   & other  & 460   & 0.0035 \\
X     & other  & 916   & 0.0070 \\
\bottomrule
\end{tabular}
\caption{Non-zero latents and fraction of latent space by POS tag.}
\label{tab:nonzero-pos-latents}
\end{table}

\begin{table}[t]
\footnotesize
\centering
\begin{tabular}{lrr}
\toprule
\textbf{\textsc{PoS} Class} & \textbf{mean$\pm$ std.} \\
\midrule
Closed & 3281.74 $\pm$ 1895.65 \\
Open & 7926.52 $\pm$ 2153.69 \\
Other & 2094.42 $\pm$ 221.91 \\
\bottomrule
\end{tabular}
\caption{Number of latents with non-zero coefficients for classification for each one-vs-rest classifier, aggregated by \textsc{PoS} Group. Classifiers for \textsc{Open}-class \textsc{PoS} tags have the highest number of non-zero coefficients.}\label{tab:feat_salience}
\end{table}

\paragraph{Coverage and Compactness.}  

In Table \ref{tab:coverage-on-active} we report, for each \textsc{PoS}, the ratio between the coverage for $L^{(k^\star_{c'})}$ and the number of latents with non-zero coefficients for $c'$. Intuitively, this represent the proportion of latents needed to reach coverage for a \textsc{PoS} with respect to all salient latents for the \textsc{PoS}.

\begin{table}[t]
\footnotesize
\centering
\begin{tabular}{llc}
\toprule
\textbf{Class} & \textbf{POS} & Coverage over Salience (\%) \\
\midrule
\multirow{6}{*}{Open}
 & \textsc{adj}   & 0.91 \\
 & \textsc{adv}   & 1.38 \\
 & \textsc{intj}  & 3.76 \\
 & \textsc{noun}  & 0.34 \\
 & \textsc{propn} & 1.14 \\
 & \textsc{verb}  & 0.70 \\
\midrule
\multirow{8}{*}{Closed}
 & \textsc{adp}   & 0.25 \\
 & \textsc{aux}   & 1.06 \\
 & \textsc{cconj} & 0.62 \\
 & \textsc{det}   & 0.42 \\
 & \textsc{num}   & 1.20 \\
 & \textsc{part}  & 0.47 \\
 & \textsc{pron}  & 0.76 \\
 & \textsc{sconj} & 0.45 \\
\midrule
\multirow{3}{*}{Other}
 & \textsc{punct} & 0.33 \\
 & \textsc{sym}   & 5.65 \\
 & \textsc{x}     & 8.95 \\
\bottomrule
\end{tabular}
\caption{Coverage over number of salient features per UPOS, grouped by class.}
\label{tab:coverage-on-active}
\end{table}

\paragraph{Compact-feature classification.} 
Table \ref{tab:multiclass-classifier-results} reports per-\textsc{PoS} performances of the multiclass classifier trained with the compact feature set $L^*$.

\begin{table}[t]
    \scriptsize
    \centering

\begin{tabular}{llllr}
\toprule
 & precision & recall & f1-score & support \\
\midrule
ADJ & 0.85 & 0.84 & 0.85 & 11489 \\
ADP & 0.93 & 0.88 & 0.91 & 16655 \\
ADV & 0.84 & 0.79 & 0.81 & 8556 \\
AUX & 0.93 & 0.91 & 0.92 & 9682 \\
CCONJ & 0.96 & 0.97 & 0.96 & 5854 \\
DET & 0.95 & 0.93 & 0.94 & 14307 \\
INTJ & 0.36 & 0.89 & 0.52 & 1860 \\
NOUN & 0.92 & 0.87 & 0.90 & 29289 \\
NUM & 0.88 & 0.93 & 0.90 & 3368 \\
PART & 0.82 & 0.95 & 0.88 & 4314 \\
PRON & 0.95 & 0.93 & 0.94 & 15197 \\
PROPN & 0.83 & 0.83 & 0.83 & 10144 \\
PUNCT & 0.99 & 0.94 & 0.96 & 24563 \\
SCONJ & 0.66 & 0.86 & 0.75 & 2878 \\
SYM & 0.33 & 0.96 & 0.49 & 281 \\
VERB & 0.94 & 0.86 & 0.90 & 18642 \\
X & 0.14 & 0.88 & 0.24 & 331 \\ 
\midrule
accuracy & & & 0.89 & 177410\\
macro avg & 0.78 & 0.89 & 0.81 & 177410 \\
weighted avg & 0.91 & 0.89 & 0.90 & 177410 \\
\bottomrule
\end{tabular}
\caption{Multiclass classifier performances.}\label{tab:multiclass-classifier-results}
\end{table}

\begin{figure*}[t]
    \centering
    \includegraphics[width=\linewidth]{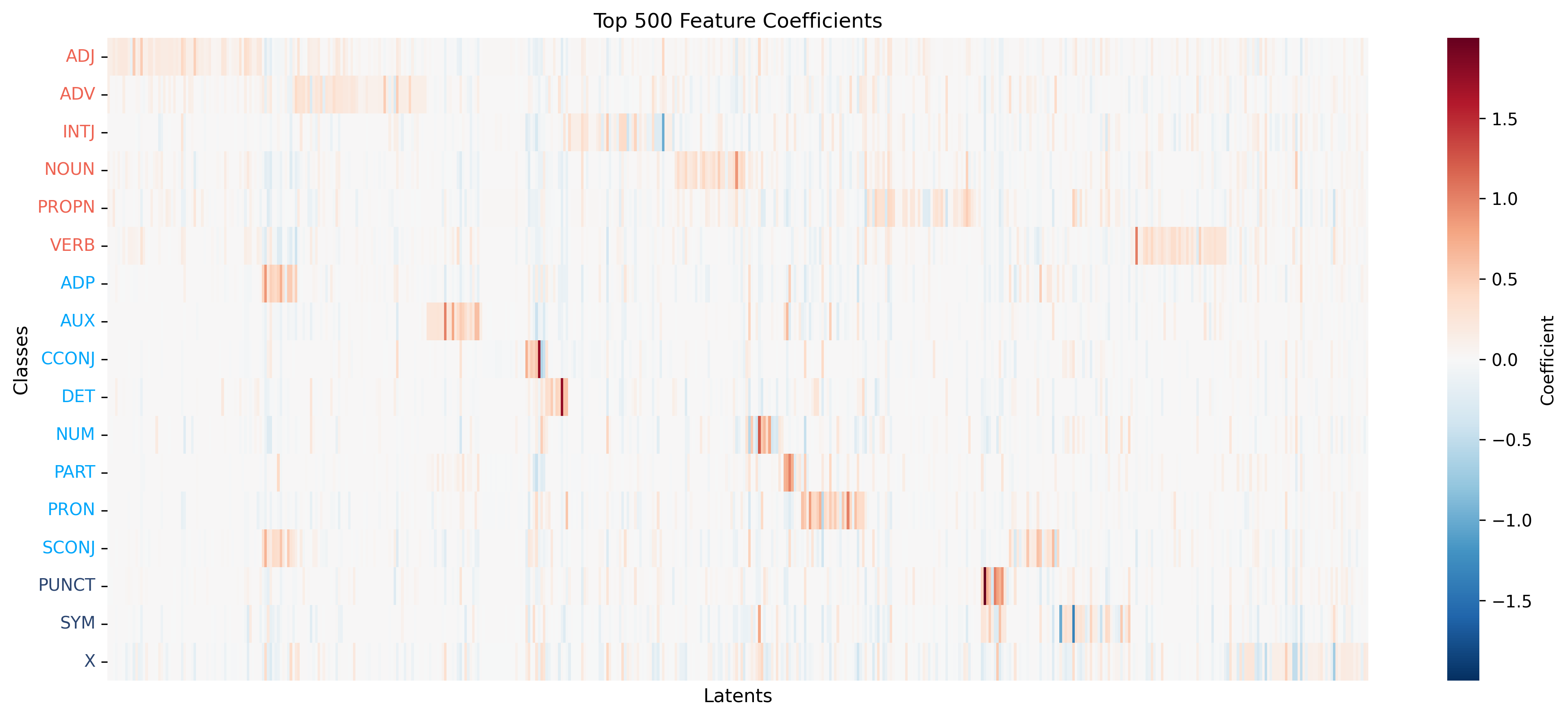}
    \caption{Coefficients importance heatmap for each \textsc{PoS} in the mutliclass classifier trained with 5-fold cross-validation on the GUM Training set.}
    \label{fig:coeff-heatmap}

\end{figure*}

Figure \ref{fig:coeff-heatmap} shows the heatmap of coefficient association to each \textsc{PoS} class in the multilabel classification experiment. Coefficient are sorted for importance across \textsc{PoS} classes.

\begin{figure*}[t]
    \centering
    \includegraphics[width=\linewidth]{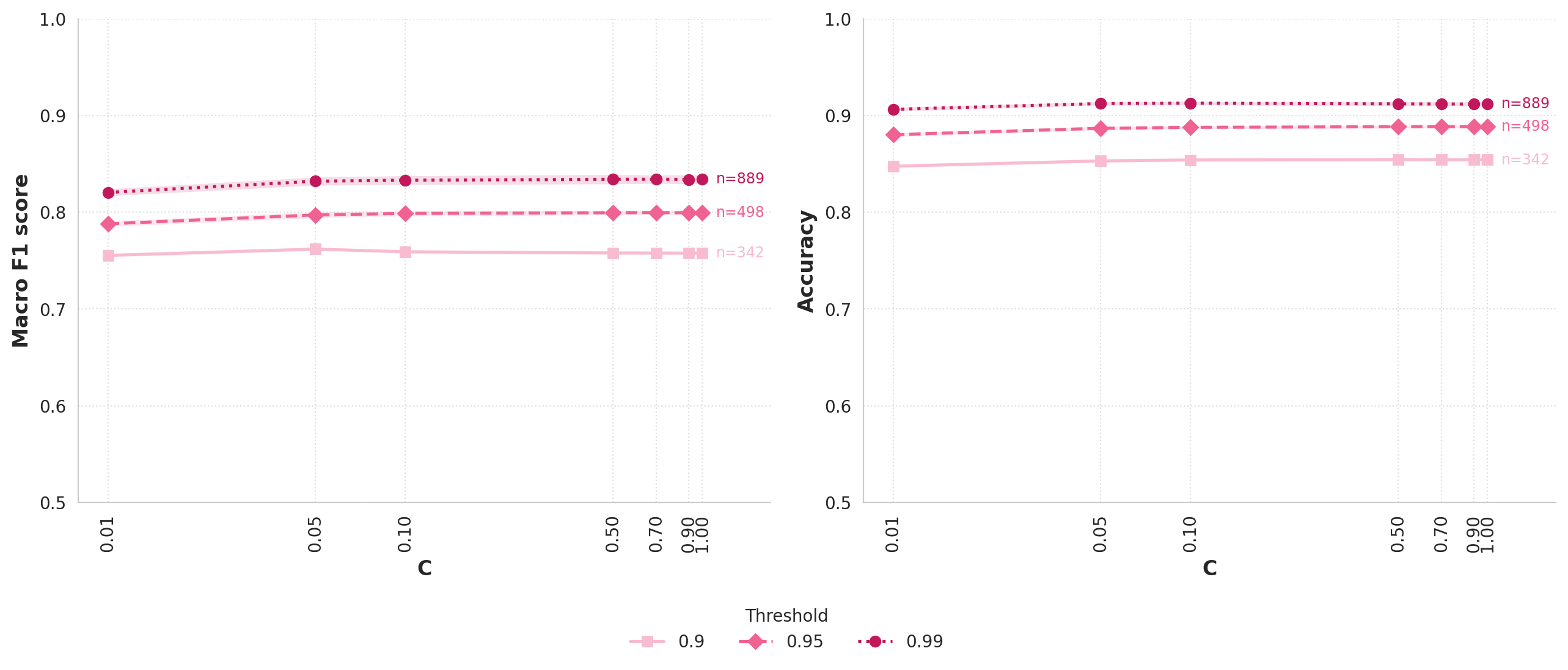}
    \caption{Parameter sweep for $C$ and $\tau$ for the compact feature classification task.}
    \label{fig:sweep-c-tau}

\end{figure*}

Figure \ref{fig:sweep-c-tau} provides a sensitivity analysis of the performances of the classifier with respect to values of $C$ and $\tau$. We report mean Macro-F1 score and Accuracy in the cross validation setting. We also provide standard deviation in the form of error bars. From the plot, it clearly emerges that the classification results are not particularly sensitive neither to the $\tau$ threshold nor the $C$ value. As for the $\tau$, performances increase monotonically, but with a difference of $\sim$5 points using 3x less features. As for the $C$ values, performances remain almost identical, and the standard deviation is near zero, indicating no meaningful differences.

\subsection{RQ3: Validation on held-out data}\label{app:heldout-data}

\begin{figure}[t]
    \centering
    \includegraphics[width=\linewidth]{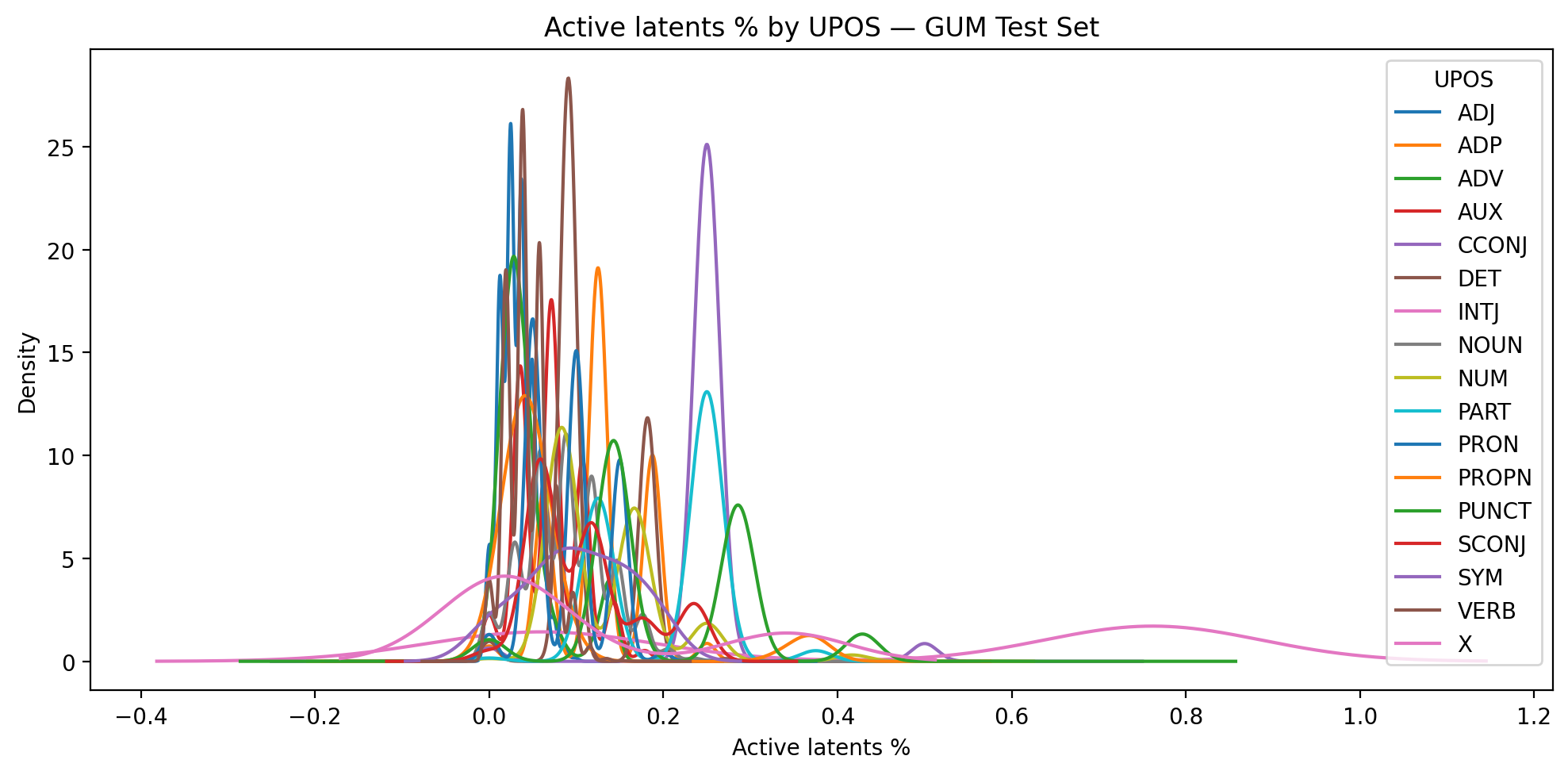}
    \caption{Density plot: number of activations from $L^{(k^\star_{c'})}$ over \textsc{PoS} with tag $c'$ in the Treebank test set, for all $c' \in C$.}
    \label{fig:active-latents-treebank}

\end{figure}

\paragraph{Density of Activations per \textsc{PoS}.}\label{par:pos-density-treebank-test} In Figure \ref{fig:active-latents-treebank} we report a KDE plot representing density of number of activations from $L^{(k^\star_{c'})}$ over \textsc{PoS} with tag $c'$ in the Treebank test set, for all $c' \in C$. The Figure shows that in most cases at least 1 to 4\% of latents in $L^{(k^\star_{c'})}$ fire on all tokens of $c'$.

We further provide indications that most tokens associated with category $c'$ fire latents in $L^{(k^\star_{c'})}$. Over the whole treebank test set, we count the number of cases in which no latents in $L^{(k^\star_{c'})}$ fire over a token of category $c'$. Table \ref{tab:combined-metrics} shows the fraction of tokens with no  $L^{(k^\star_{c'})}$ for each $c' \in C$.

\paragraph{Distinctiveness.}\label{app:distinctiveness}
Table \ref{tab:combined-metrics} reports per-$c$ results of $M_{c,c}$ and distinctiveness $D(c)$.
We observe that \textsc{INTJ} and \textsc{X} have the least distinctive activations. In this case we do not see a clear \textsc{PoS}-class based trend with regards to distinctiveness.

\begin{table}[t]
\footnotesize
\centering
\begin{tabular}{llcrr}
\toprule
\textbf{Class} & \textbf{\textsc{POS}} & \textbf{Zero-act.\ \%} & \textbf{$M_{cc}$} & \textbf{$D(c)$} \\
\midrule
\multirow{6}{*}{Open}
 & \textsc{adj}   & 0.056 & 0.944 & 0.305 \\
 & \textsc{adv}   & 0.040 & 0.960 & 0.217 \\
 & \textsc{intj}  & 0.076 & 0.924 & 0.093 \\
 & \textsc{noun}  & 0.051 & 0.949 & 0.336 \\
 & \textsc{propn} & 0.027 & 0.973 & 0.273 \\
 & \textsc{verb}  & 0.048 & 0.952 & 0.303 \\
\midrule
\multirow{8}{*}{Closed}
 & \textsc{adp}   & 0.019 & 0.981 & 0.350 \\
 & \textsc{aux}   & 0.048 & 0.952 & 0.301 \\
 & \textsc{cconj} & 0.024 & 0.976 & 0.261 \\
 & \textsc{det}   & 0.023 & 0.977 & 0.350 \\
 & \textsc{num}   & 0.006 & 0.994 & 0.345 \\
 & \textsc{part}  & 0.007 & 0.993 & 0.317 \\
 & \textsc{pron}  & 0.030 & 0.970 & 0.225 \\
 & \textsc{sconj} & 0.024 & 0.976 & 0.224 \\
\midrule
\multirow{3}{*}{Other}
 & \textsc{punct} & 0.051 & 0.949 & 0.333 \\
 & \textsc{sym}   & 0.111 & 0.889 & 0.299 \\
 & \textsc{x}     & 0.273 & 0.727 & 0.122 \\
\bottomrule
\end{tabular}
\caption{Zero-activation fraction, $M_{cc}$, and distinctiveness $D(c)$ per UPOS category.}
\label{tab:combined-metrics}
\end{table}

\paragraph{Per \textsc{Pos} Activation distribution in the controlled dataset.}\label{app:KDE_curated_dataset}

\begin{figure}[t]     
\centering \includegraphics[width=0.9\columnwidth, trim=40pt 8pt 10pt 60pt, clip]{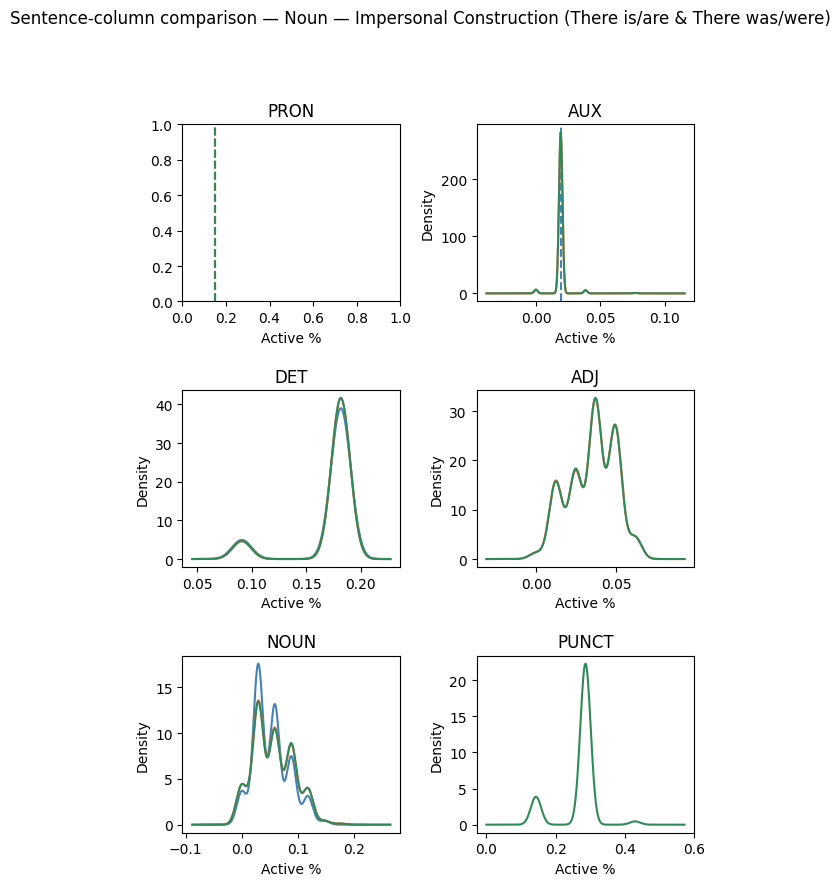}
    \caption{KDE of active-latent percentages per \textsc{PoS} tag in the \textit{I} + [VERB] templates. 
    }
    \label{fig:sent_col_comparison_fig7}
\end{figure}

\begin{figure}[t]     
\centering \includegraphics[width=0.65\columnwidth, trim=70pt 8pt 30pt 75pt, clip]{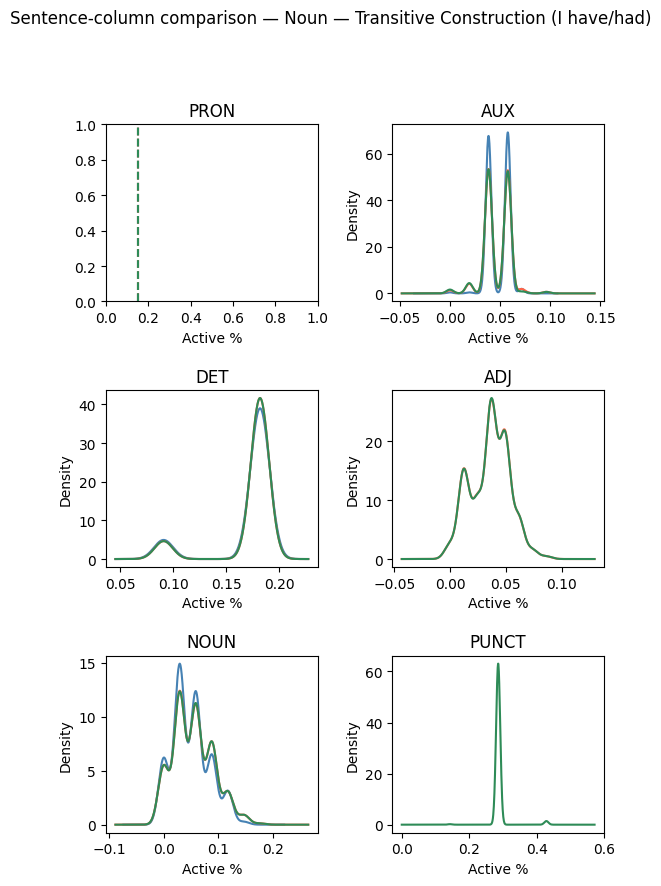}
    \caption{KDE of active-latent percentages per \textsc{PoS} tag in the \textit{I have/had} templates.
    }
    \label{fig:sent_col_comparison_I_have_had}
    
\end{figure}

\begin{figure}[t]     
\centering \includegraphics[width=0.65\columnwidth, trim=40pt 8pt 10pt 60pt, clip]{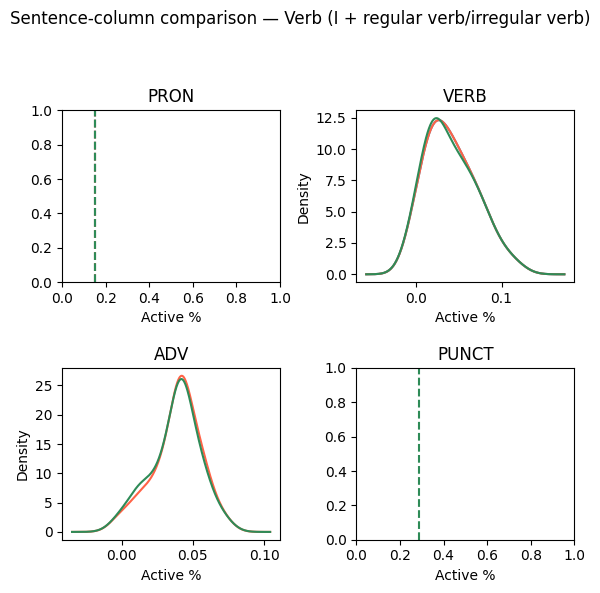}
    \caption{KDE of active-latent percentages per \textsc{PoS} tag in the \textit{I} + [VERB] templates. 
    }
    \label{fig:sent_col_comparison_verb}
\end{figure}

\begin{figure}[t]     
\centering \includegraphics[width=0.65\columnwidth, trim=50pt 8pt 30pt 75pt, clip]{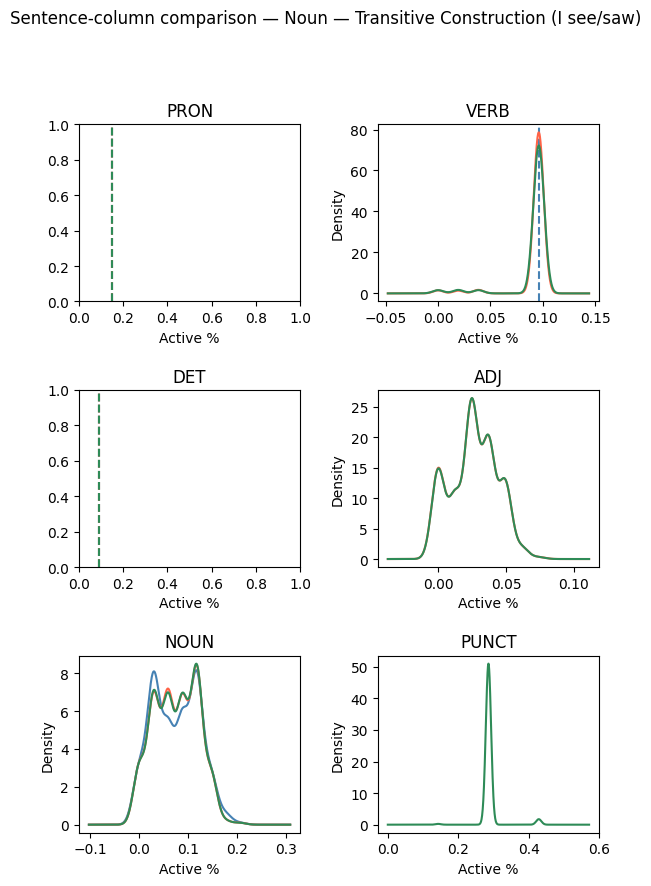}
    \caption{KDE of active-latent percentages per \textsc{PoS} tag in the \textit{There is/are} templates. 
    }
    \label{fig:sent_col_comparison_I_see_saw}
\end{figure}

Figures \ref{fig:sent_col_comparison_fig7} through \ref{fig:sent_col_comparison_I_see_saw} report the KDE distributions of active-latent percentages per \textsc{PoS} tag for the remaining three construction types in the controlled dataset. In each figure, \textcolor{steelblue}{\textbf{blue}}, \textcolor{tomato}{\textbf{red}},
and \textcolor{seagreen}{\textbf{green}} correspond to sentences of increasing syntactic complexity: the minimal construction (e.g., [PRON] + [AUX] + [DET] + [NOUN]), the extension with an adjective \textsc{[ADJ]} or an \textsc{[ADV]} for the verb template, and the further addition of punctuation \textsc{[PUNCT]}, respectively; dashed vertical lines indicate cases where only a single value is available for a given \textsc{PoS} tag and sentence type. 

The distributions for most \textsc{PoS} tags are highly stable across sentence types, with the three curves largely overlapping. A consistent exception is the \textsc{noun} tag: in both transitive constructions (\textit{I have/had} and \textit{I see/saw}), the \textcolor{tomato}{\textbf{red}} and \textcolor{seagreen}{\textbf{green}} curves, corresponding to sentences containing an additional adjective, show a slight shift in the activation distribution relative to the \textcolor{steelblue}{\textbf{blue}} curve. This suggests that the presence of an adjacent adjective marginally affects noun-associated latent activations, consistent with co-activation patterns 
discussed in Section~\ref{sec:ling_analysis}; similarly, \textsc{aux} in 
Figure~\ref{fig:sent_col_comparison_I_have_had} displays two distinct peaks, 
reflecting the alternation between the present \textit{have} and past \textit{had} 
forms across sentences, confirming latents' sensitivity to surface form in \textsc{PoS} 
classes with relatively low token variability. By contrast, the \textsc{verb} tag in Figure~\ref{fig:sent_col_comparison_verb} 
shows only a minor shift across sentence types, suggesting that verb-associated 
latents are less sensitive to the surrounding syntactic context than noun-associated ones.

\subsubsection{First-token activations in the controlled dataset}\label{app:first-tok-act}

\begin{figure}[t]     
\centering \includegraphics[width=\columnwidth]{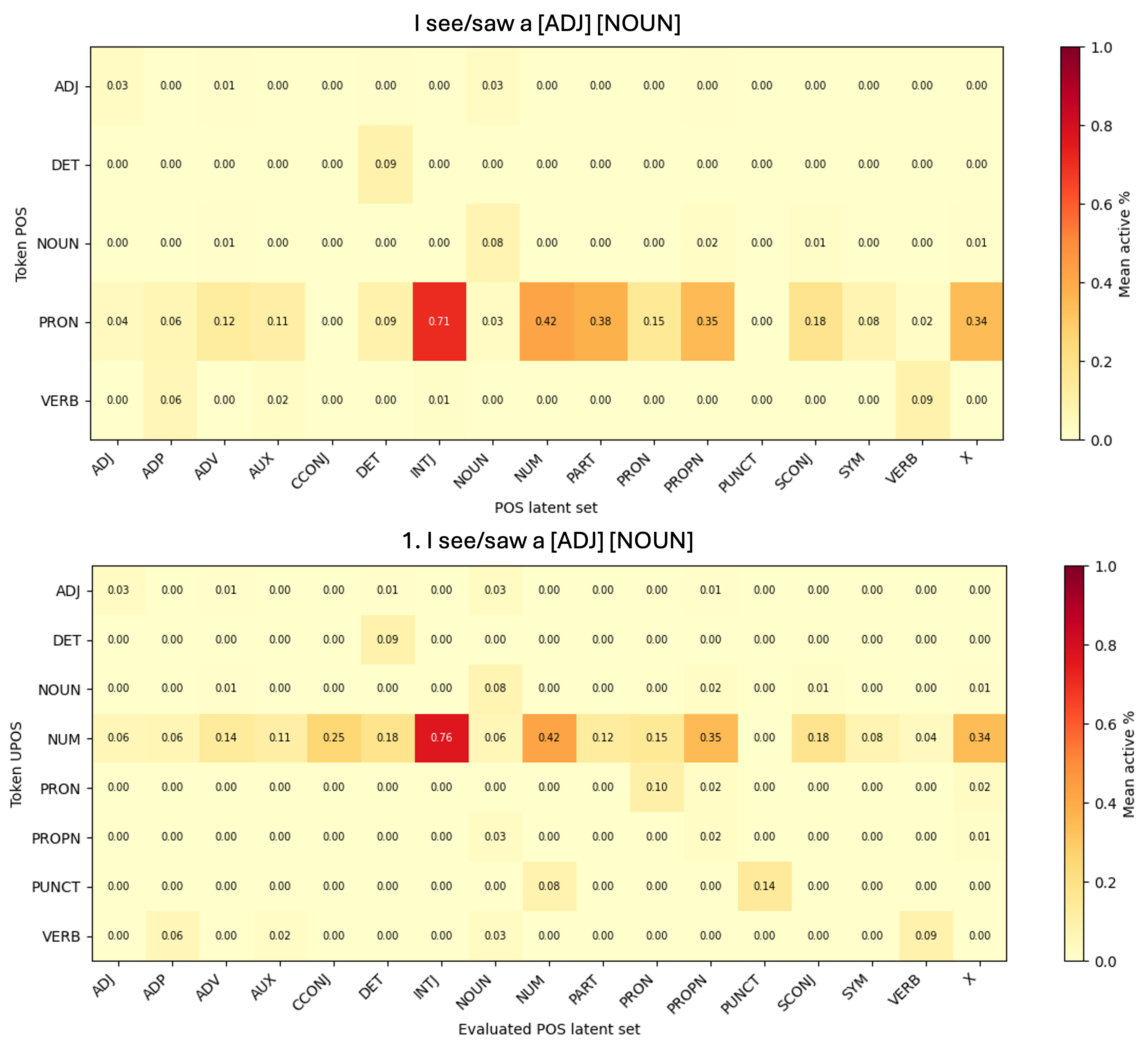}
    \caption{Demonstration of the behavior on first-token activations. We show that several activations shift from \textsc{PRON} in the top sentence (the first token ``I'') to the first token \textsc{NUM} in the bottom sentence.}
    \label{fig:first-token-demo}
\end{figure}

We report an example of first-token activations confounds on the controlled dataset. All tested cases show the same behavior, but we report only one example for brevity.
We do the following: we prepend ``1.'' to all templates, e.g., ``1. I saw the cute cat'', recompute activations, and compare pre-vs-post addition of the \textsc{NUM}+\textsc{PUNCT}. In Figure \ref{fig:first-token-demo} we report the results for the ``I see/saw a [ADJ] [NOUN]''. We observe that all latents that fire on \textsc{PRON} in the original sentence (the first token ``I'') shift to the first token \textsc{NUM}.

\end{document}